%% file: main_camera_ready.tex
\documentclass{article} % For LaTeX2e
\usepackage{iclr2021_conference,times}

\usepackage[utf8]{inputenc} % allow utf-8 input
\usepackage[T1]{fontenc}    % use 8-bit T1 fonts
\usepackage{hyperref}       % hyperlinks
\usepackage{url}            % simple URL typesetting
\usepackage{booktabs}       % professional-quality tables
\usepackage{amsfonts}       % blackboard math symbols
\usepackage{nicefrac}       % compact symbols for 1/2, etc.
\usepackage{microtype}      % microtypography
\usepackage{amsmath}
\usepackage{amssymb}
\usepackage{sidecap}

\usepackage{graphicx, wrapfig}
\usepackage[font=small]{caption}
\usepackage[font=small]{subcaption}
\usepackage[table]{xcolor}
\usepackage{mathtools}
\usepackage{wrapfig}
\usepackage{multirow}
\usepackage{enumitem}
\usepackage{xspace}
\usepackage{authblk}

\newcommand{\arch}{\texttt{DeLighT\xspace}} %tranformation function name
\newcommand{\dextra}{\texttt{DeLighT} transformation\xspace} % attention unit name
\newcommand{\sota}{state-of-the-art}

\usepackage{tikz}
\usetikzlibrary{decorations.pathreplacing, calligraphy}
\usetikzlibrary{shapes.callouts}
\usetikzlibrary{chains, fit, quotes, automata}
\usetikzlibrary{arrows}
\usetikzlibrary{matrix,positioning}
\usetikzlibrary{shapes.geometric, shapes.misc}
\usetikzlibrary{calc}

\usepackage{listings}

\DeclarePairedDelimiter\ceil{\lceil}{\rceil}

\title{DeLighT: Deep and Light-weight Transformer}

\makeatletter
\renewcommand\AB@affilsepx{\quad\protect\Affilfont}
\makeatother

\author[1]{Sachin Mehta}
\author[2]{Marjan Ghazvininejad}
\author[2]{Srinivasan Iyer}
\author[1, 2]{\\ Luke Zettlemoyer} 
\author[1,3]{Hannaneh Hajishirzi}

\affil[1]{University of Washington}
\affil[2]{Facebook AI Research}
\affil[3]{Allen Institute for AI}

\iclrfinalcopy

\begin{document}

\maketitle

\begin{abstract}
We introduce a deep and light-weight transformer, \arch, that delivers similar or better performance than standard transformer-based models with significantly fewer parameters. \arch~more efficiently allocates parameters both (1) within each Transformer block using the \dextra, a deep and light-weight transformation and (2) across blocks using block-wise scaling, that allows for shallower and narrower \arch~blocks near the input and wider and deeper \arch~blocks near the output. Overall, \arch~networks are 2.5 to 4 times deeper than standard transformer models and yet have fewer parameters and operations. Experiments on benchmark machine translation and language modeling tasks show that \arch~matches or improves the performance of baseline Transformers with 2 to 3 times fewer parameters on average. 
\end{abstract}

\section{Introduction}
Attention-based transformer networks~\citep{vaswani2017attention} are widely used for sequence modeling tasks, including language modeling and machine translation. To improve performance, models are often scaled to be either wider, by increasing the dimension of hidden layers, or deeper, by stacking more transformer blocks. For example, T5 \citep{raffel2019exploring} uses a dimension of 65K and GPT-3 \citep{brown2020language} uses 96 transformer blocks. However, such scaling increases the number of network parameters significantly (e.g., T5 and GPT-3 have 11 billion and 175 billion parameters, respectively), and complicates learning, i.e., these models either require very large training corpora \citep{raffel2019exploring,devlin2018bert,brown2020language} or careful regularization \citep{hinton2012improving,wan2013regularization,merity2018regularizing}. In this paper, we introduce a new parameter-efficient attention-based architecture that can be easily scaled to be both wide and deep. 

Our \texttt{De}ep and \texttt{Ligh}t-weight \texttt{T}ransformer architecture, \arch, extends the transformer architecture of \citet{vaswani2017attention} and delivers similar or better performance with significantly fewer parameters and operations. At the heart of \arch~is the \dextra~that uses the group linear transformations (GLTs) of \citet{mehta2018pyramidal} with an expand-reduce strategy for varying the width and depth of the \arch~block efficiently. Since GLTs are local by nature, the \dextra~uses feature shuffling, which is analogous to channel shuffling in convolutional networks \citep{zhang2018shufflenet}, to share information between different groups. Such wide and deep representations facilitate replacing the multi-head attention and feed-forward layers in transformers with single headed attention and light-weight feed-forward layers, reducing total network parameters and operations. Importantly, unlike transformers, the \dextra~decouples the depth and width from the input size, allowing us to allocate parameters more efficiently across blocks by using shallower and narrower \arch~blocks near the input and deeper and wider \arch~blocks near the output.

We demonstrate that \arch~models achieve similar or better performance than transformer models with significantly fewer parameters and operations, on two common sequence modeling tasks, (i) machine translation and (ii)  language modeling. On the low resource WMT'16 En-Ro machine translation dataset, \arch~attains transformer performance using $2.8\times$ fewer parameters. On the high resource WMT'14 En-Fr dataset, \arch~delivers better performance (+0.4 BLEU score) with $1.8\times$ fewer parameters than baseline transformers. Similarly, on language modeling, \arch~matches the performance of  Transformer-XL~\citep{dai2019transformer} with $1.5\times$ fewer parameters on the WikiText-103 dataset. Our source code is open-source and is available at: \textcolor{blue}{\url{https://github.com/sacmehta/delight}}

\section{Related Work}
\label{sec:related_work}

\noindent{\bf Improving transformers:} Several methods have been introduced to improve the transformer architecture. The first line of research addresses the challenge of computing self attention on long input sequences \citep{child2019generating,Kitaev2020Reformer,Beltagy2020Longformer}. These methods can be combined with our architecture. The second line of research focuses on explaining multi-head attention \citep{raganato2018analysis,Brunner2020On}. They show that increasing the number of transformer heads can lead to redundant representations \citep{voita2019bottom,michel2019sixteen} and using fixed attention heads with predefined patterns \citep{raganato2020fixed} or synthetic attention matrices \citep{tay2020synthesizer} improves performance. 
%These results support our design choice of using single-head attention. 
The third line of research focuses on improving transformers by learning better representations \citep{wu2018pay,Wu2020Lite,so2019evolved}. These works aim to improve the expressiveness of transformers using different transformations -- for example, using convolutions \citep{wu2018pay, gehring2017convolutional}, gated linear units \citep{dauphin2017language}, or multi-branch feature extractors \citep{so2019evolved,Wu2020Lite}. Our work falls into this category. Unlike previous works, we show that it is possible to efficiently allocate parameters both at the block-level using the \dextra~and across blocks using block-wise scaling.

\noindent{\bf Model scaling:} Model scaling is a standard method to improve the performance of sequence models \citep{vaswani2017attention,raffel2019exploring,lan2020ALBERT,devlin2018bert,shoeybi2019megatron,tan2019efficientnet,brown2020language}. Model dimensions are increased in width-wise scaling \citep{vaswani2017attention,devlin2018bert} while more blocks (e.g., Transformer blocks) are stacked in depth-wise scaling \citep{shoeybi2019megatron,brown2020language,wang2019learning}. In both cases (and their combination), parameters inside each block of the network are the same, which may lead to a sub-optimal solution. To further improve the performance of sequence models, this paper introduces \textit{block-wise scaling} that allows for variably-sized blocks and efficient allocation of parameters in the network. Our results show that (1) shallower and narrower \arch~blocks near the input and deeper and wider \arch~blocks near the output deliver the best performance, and (2) models with block-wise scaling coupled with model scaling achieve better performance compared to model scaling alone. We note that convolutional neural networks (CNNs) also learn shallower and narrower representations near the input and deeper and wider representations near the output. Unlike CNNs (e.g., ResNet of \citealt{he2016deep}) that perform a fixed number of operations at each convolutional layer, the proposed block-wise scaling uses a variable number of operations in each layer and block. 

\noindent{\bf Improving sequence models:} There is also significant recent work on other related methods for improving sequence models, including (1) improving accuracy using better token-level representations -- for example, using BPE \citep{sennrich2015neural}, adaptive inputs \citep{baevski2018adaptive} and outputs \citep{grave2017efficient}, and DeFINE \citep{mehta2020DeFINE}, and (2) improving efficiency -- for example, using compression \citep{chen2018groupreduce,sun2020mobilebert}, pruning \citep{han2015deep,voita2019analyzing}, and distillation \citep{hinton2015distilling,sanh2019distilbert}. The closest to our work is the DeFINE transformation, which also learns representations using an expand-reduce strategy. The key difference between the DeFINE transformation (Figure \ref{fig:hgt}) and the \dextra~(Figure \ref{fig:ihgt}) is that the \dextra~more efficiently allocates parameters within expansion and reduction layers. Unlike DeFINE, which uses fewer groups in group linear transformations to learn wider representations, \dextra~uses more groups to learn wider representations with fewer parameters. The \dextra~achieves comparable performance to the DeFINE transformation but with significantly fewer parameters.

\section{DeLighT: Deep and Light-weight Transformer}
\label{sec:arcitecture}

\begin{figure}[t!]
    \centering
    \begin{subfigure}[b]{0.35\columnwidth}
        \centering
        \resizebox{!}{160px}{
            \input{tikz/self_attention_units.tikz}\transformer
        }
        \caption{Transformer block}
        \label{fig:transformer_sa}
    \end{subfigure}
    \hfill
    \begin{subfigure}[b]{0.33\columnwidth}
        \centering
        \resizebox{!}{165px}{
            \input{tikz/self_attention_units.tikz}\redefine
        }
        \caption{\arch~ block}
        \label{fig:redefine_transformer_sa}
    \end{subfigure}
    \hfill
    \begin{subfigure}[b]{0.3\columnwidth}
        \centering
        \begin{subfigure}[b]{\columnwidth}
            \centering
            \resizebox{!}{70px}{
                \input{tikz/dextra_new}\define
            }
            \caption{DeFINE transformation}
            \label{fig:hgt}
            \end{subfigure}
            \vfill
            \begin{subfigure}[b]{\columnwidth}
            \centering
            \resizebox{!}{70px}{
                \input{tikz/dextra_new}\invHGTShuffle
            }
            \caption{\dextra}
            \label{fig:ihgt}
        \end{subfigure}
    \end{subfigure}
    \caption{\textbf{(a, b)} Block-wise comparison between the standard transformer block of \citet{vaswani2017attention} and the \arch~block. In the \dextra, the number of operations in computing attention are reduced by half while the number of parameters (and operations) in the FFN are reduced by $16\times$. Transformations with learnable parameters (\colorbox{cyan!40}{Linear} and \colorbox{green!20}{\arch}) are shown in color. The shape of linear transformations indicate their operation (expansion, reduction, etc.). \textbf{(c, d)} compares the DeFINE transformation \citep{mehta2020DeFINE} with the \dextra. Compared to the DeFINE transformation, the \dextra~uses group linear transformations (GLTs) with more groups to learn wider representations with fewer parameters. Different colors are used to show groups in GLTs. For simplicity, feature shuffling is not shown in (d).}
    \label{fig:compare_tam_ram_sa}
\end{figure}
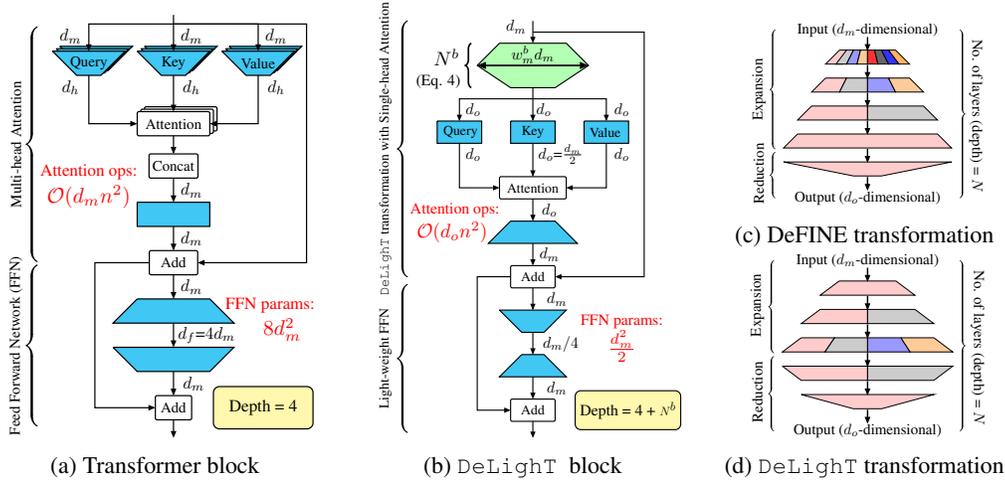

A standard transformer block (Figure \ref{fig:transformer_sa}) comprises of multi-head attention that uses a query-key-value decomposition to model relationships between sequence tokens, and a feed forward network (FFN) to learn wider representations. Multi-head attention obtains query $\mathbf{Q}$, key $\mathbf{K}$, and value $\mathbf{V}$ by applying three projections to the input, each consisting of $h$ linear layers (or heads) that map the $d_m$-dimensional input into a $d_h$-dimensional space, where $d_h=d_m/h$ is the head dimension. The FFN consists of two linear layers, where the first expands the dimensions from $d_m$ to $d_f$ and the second reduces the dimensions from $d_f$ to $d_m$. The depth of a transformer block is 4, consisting of (1) three parallel branches for queries, keys, and values, (2) a fusion layer that combines the output of multiple heads, and (3) two sequential linear layers in the FFN. In general, transformer-based networks sequentially stacks transformer blocks to increase network capacity and depth.

This paper extends the transformer architecture and introduces a deep and light-weight transformer, \arch. Our model uses a deep and light-weight expand-reduce transformation, \dextra~(Section \ref{ssec:dextra}), that enables learning wider representations efficiently. It also enables replacing multi-head attention and feed forward network (FFN) layers with single-head attention and a light-weight FFN (Section \ref{ssec:dextra_transformer}). \dextra~decouples attention dimensions from the depth and width, allowing us to learn representations efficiently using block-wise scaling instead of uniform stacking of transformer blocks (Section \ref{ssec:layer_wise_scaling}).
\subsection{DeLighT Transformation}
\label{ssec:dextra}
\dextra~maps a  $d_m$ dimensional input vector into a high dimensional space (expansion) and then reduces it down to a  $d_o$ dimensional output vector (reduction) using  $N$ layers of the group transformations of \citet{mehta2018pyramidal}, as shown in Figure \ref{fig:ihgt}. During these expansion and reduction phases, \dextra~uses group linear transformations (GLTs) because they learn local representations by deriving the output from a specific part of the input and are more efficient than linear transformations. To learn global representations, the \dextra~shares information between different groups in the group linear transformation using feature shuffling, analogous to channel shuffling in convolutional networks \citep{zhang2018shufflenet}. 

A standard approach to increase the expressivity and capacity of transformers is to increase the input dimensions, $d_m$. However, increasing $d_m$ linearly also increases the number of operations in multi-head attention ($\mathcal{O}(n^2 d_m)$, where $n$ is the sequence length) in a standard transformer block (Figure \ref{fig:transformer_sa}). In contrast, to increase the expressivity and capacity of the \arch~block, we increase the depth and width of its intermediate \arch~ transformations using expansion and reduction phases. This enables us to use smaller dimensions for computing attention, requiring fewer operations. 

\begin{figure}[t!]
    \centering
    \includegraphics[height=110px]{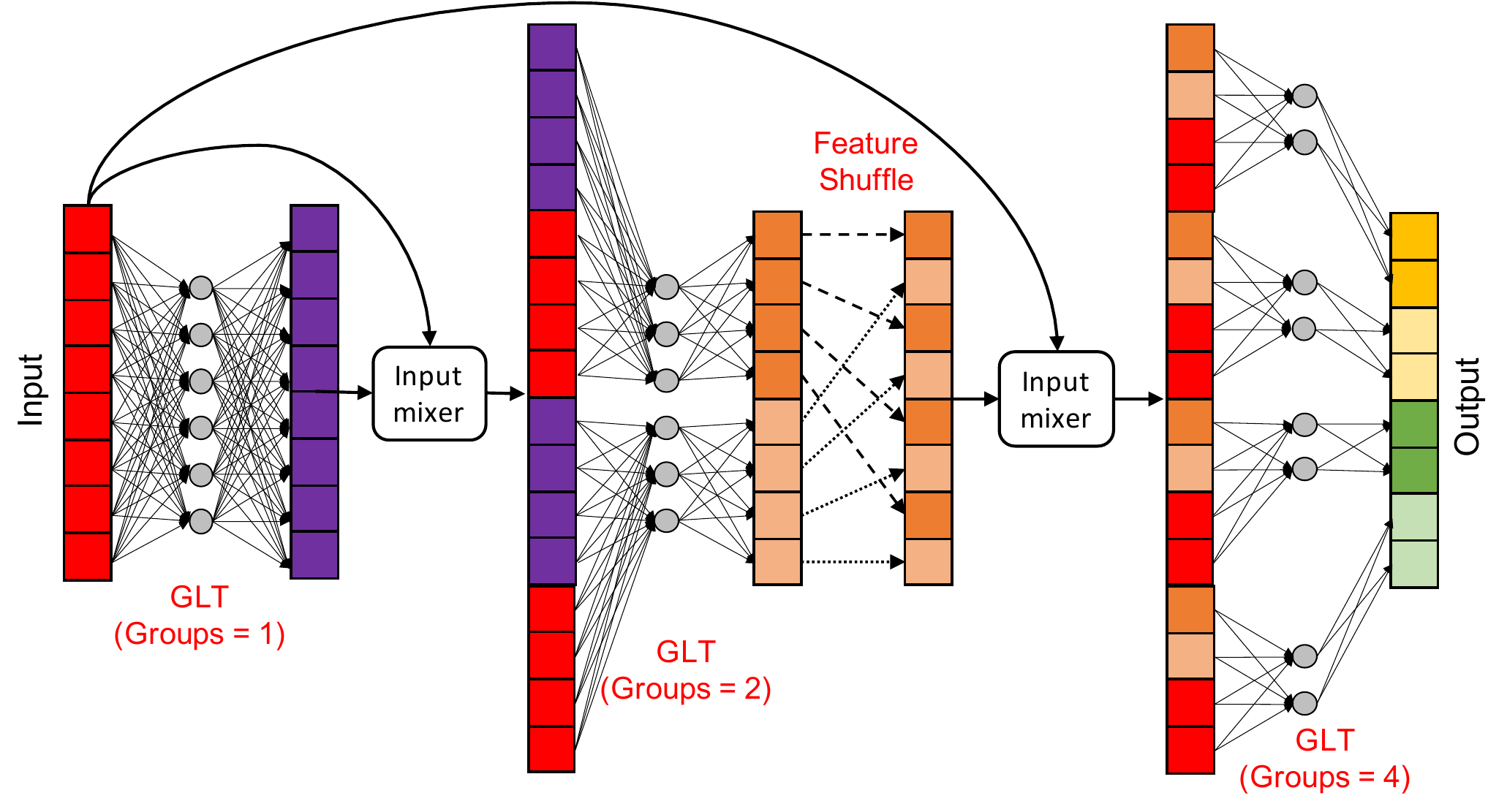}
    \caption{Example illustrating the expansion phase in the \arch~transformation that uses GLTs, feature shuffling, and an input mixer connection, to learn deeper and wider representations efficiently. For illustrative purposes, we have used the same input and output dimensions.}
    \label{fig:delight_layer_vis}
\end{figure}

Formally, the \dextra~is controlled by five configuration parameters: (1) number of GLT layers $N$, (2) width multiplier $w_m$, (3) input dimension $d_m$, (4) output dimension $d_o$, and (5) maximum groups $g_{max}$ in a GLT. In the expansion phase, the \dextra~projects the $d_m$-dimensional input to a high-dimensional space, $d_{max}=w_m d_m$, linearly using $\lceil \frac{N}{2} \rceil$ layers. In the reduction phase, the \dextra~projects the $d_{max}$-dimensional vector to a $d_o$-dimensional space using the remaining $N - \lceil \frac{N}{2} \rceil$ GLT layers. Mathematically, we  define the output $\mathbf{Y}$ at each GLT layer $l$ as:
\begin{equation}
    \mathbf{Y}^l = \left\{
    \begin{array}{ll}
        \mathcal{F} \left(\mathbf{X}, \mathbf{W}^l, \mathbf{b}^l, g^l\right), & l = 1 \\
        \mathcal{F} \left( \mathcal{H}\left(\mathbf{X}, \mathbf{Y}^{l-1}\right), \mathbf{W}^l,  \mathbf{b}^l, g^l\right), &  \text{Otherwise}
    \end{array}
    \right.
    \label{eq:glt}
\end{equation}
where $\mathbf{W}^l = \left\{ \mathbf{W}^l_1, \cdots, \mathbf{W}^l_{g^l}\right\}$ and $\mathbf{b}^l = \left\{ \mathbf{b}^l_1, \cdots, \mathbf{b}^l_{g^l}\right\}$ are the learnable weights and biases of group linear transformation $\mathcal{F}$ with $g^l$ groups at the $l$-th layer. Briefly, the $\mathcal{F}$ function takes the input $\mathbf{X}$ $\left(\text{or } \mathcal{H}\left(\mathbf{X}, \mathbf{Y}^{l-1}\right)\right)$ and splits into $g^l$ non-overlapping groups such that $\mathbf{X} = \left\{\mathbf{X}_1, \cdots, \mathbf{X}_{g^l} \right\}$. The function $\mathcal{F}$ then linearly transforms each $\mathbf{X}_i$ with weights $\mathbf{W}^l_i$ and bias $\mathbf{b}^l_i$ to produce output $\mathbf{Y}^l_i = \mathbf{X}_i \mathbf{W}_i^l + \mathbf{b}_i^l$. The outputs of each group $\mathbf{Y}^l_i$ are then concatenated to produce the output $\mathbf{Y}^l$. The function $\mathcal{H}$ first shuffles the output of each group in $\mathbf{Y}^{l-1}$ and then combines it with the input $\mathbf{X}$ using the input mixer connection of \citet{mehta2020DeFINE} to avoid vanishing gradient problems. Figure \ref{fig:delight_layer_vis} visualizes the expansion phase in the \arch~transformation with group linear transformation, feature shuffling, and the input mixer connection.

The number of groups at the $l$-th GLT in \dextra~are computed as:
\begin{equation}
    g^l = \left\{
    \begin{array}{lr}
       \text{min}(2^{l-1}, g_{max}), & 1 \leq l \leq \ceil*{N/2} \\
       g^{N-l}, & \text{Otherwise}
    \end{array}
    \right.
    \label{eq:group}
\end{equation}
In our experiments, we use $g_{max} = \lceil \frac{d_m}{32} \rceil$ so that each group has at least 32 input elements. 

\subsection{DeLighT block}  
\label{ssec:dextra_transformer} 
Figure \ref{fig:redefine_transformer_sa} shows how we integrate \dextra~into the transformer block to improve its efficiency. The $d_m$-dimensional inputs are first fed to the \dextra~to  produce $d_o$-dimensional outputs, where $d_o < d_m$. These $d_o$-dimensional outputs are then fed into a single head attention, followed by a light-weight FFN to model their relationships. 

\noindent{\bf \arch~layer and single head attention:} Let us assume we have a sequence of $n$ input tokens, each of dimensionality $d_m$. These $n$, $d_m$-dimensional inputs are first fed to the \dextra~to produce $n$, $d_o$-dimensional outputs, where $d_o < d_m$.  These $n$, $d_o$-dimensional outputs are then projected simultaneously using three linear layers to produce $d_o$-dimensional queries $\mathbf{Q}$, keys $\mathbf{K}$, and  values $\mathbf{V}$. We then model contextual relationships between these $n$ tokens using scaled dot-product attention (Eq. \ref{eq:espda}). To enable the use of residual connections \citep{he2016deep}, the $d_o$-dimensional outputs of this attention operation are  linearly projected into a $d_m$-dimensional space.
\begin{equation}
    \text{Attention}(\mathbf{K}, \mathbf{Q}, \mathbf{V}) = \text{softmax}\left(\frac{\mathbf{Q}\mathbf{K}^T}{\sqrt{d_o}}\right) \mathbf{V}
    \label{eq:espda}
\end{equation}
We hypothesize that the ability of \arch~to learn wider representations allows us to replace multi-head attention with single-head attention. The computational costs for computing attention in the standard transformer and the \arch~block are $\mathcal{O}(d_m n^2)$ and $\mathcal{O}(d_o n^2)$ respectively, where $d_o < d_m$. Therefore, the \arch~block reduces the cost for computing attention by a factor of $d_m/d_o$. In our experiments, we used $d_o=d_m/2$, thus requiring $2\times$ fewer multiplication-addition operations as compared to the transformer architecture. 

\noindent {\textbf{Light-weight FFN:}} Similar to FFNs in transformers, this block also consists of two linear layers. Since the \arch~block has already incorporated wider representations using the \dextra, it allows us to invert the functionality of FFN layers in the transformer. The first layer reduces the dimensionality of the input from $d_m$ to $d_m/r$ while the second layer expands the dimensionality from $d_m/r$ to $d_m$, where $r$ is the reduction factor  (see Figure \ref{fig:redefine_transformer_sa}). Our light-weight FFN reduces the number of parameters and operations in the FFN by a factor of $r d_f/d_m$. In the standard transformer, the FFN dimensions are expanded by a factor of $4$.\footnote{Transformer-base uses $d_m$=512 and $d_f$=2048 while Transformer-large uses $d_m$=1024 and $d_f$=4096.} In our experiments, we used $r=4$. Thus, the light-weight FFN reduces the number of parameters in the FFN by $16\times$. 

\noindent {\textbf{Block depth:}} The \arch~block stacks (1) a \dextra~with $N$ GLTs, (2) three parallel linear layers for key, query, and value, (3) a projection layer, and  (4) two linear layers of a light-weight FFN. Thus, the depth of \arch~block is $N + 4$. Compared to the standard transformer block (depth is 4), \arch~block is deeper.

\subsection{Block-wise scaling}
\label{ssec:layer_wise_scaling}
\begin{figure}[t!]
    \centering
    \begin{subfigure}[b]{0.28\columnwidth}
        \centering
        \resizebox{!}{80px}{
            \input{tikz/scaling_new.tikz}\scaling
        }
        \caption{Uniform vs. block-wise}
        \label{fig:com_tam_ram}
    \end{subfigure}
    \hfill
    \begin{subfigure}[b]{0.7\columnwidth}
        \centering
        \begin{tabular}{cc}
            \includegraphics[height=80px]{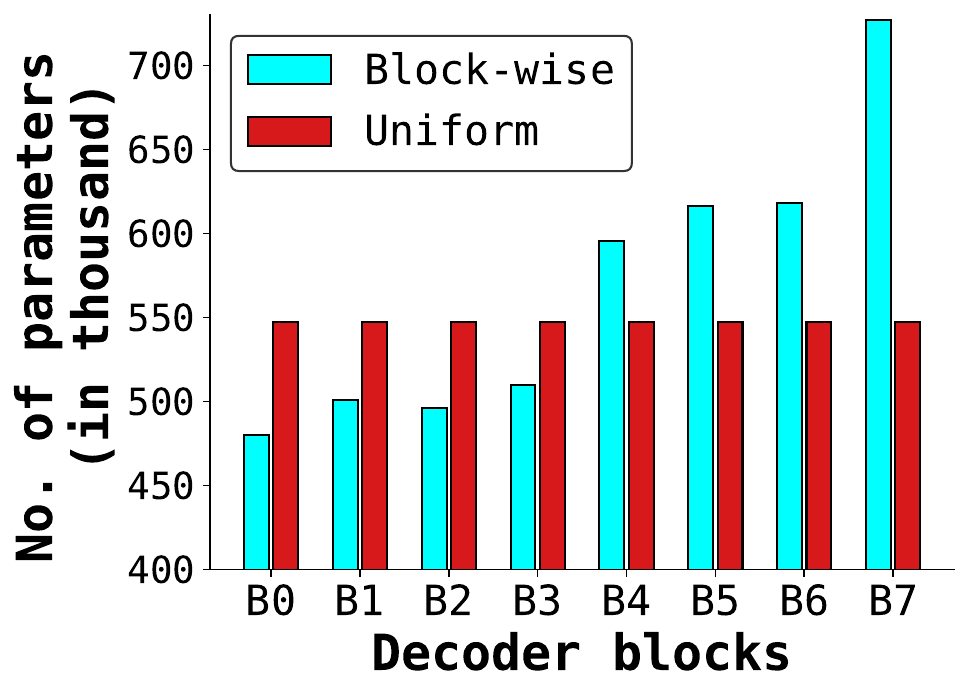} &  \includegraphics[height=80px]{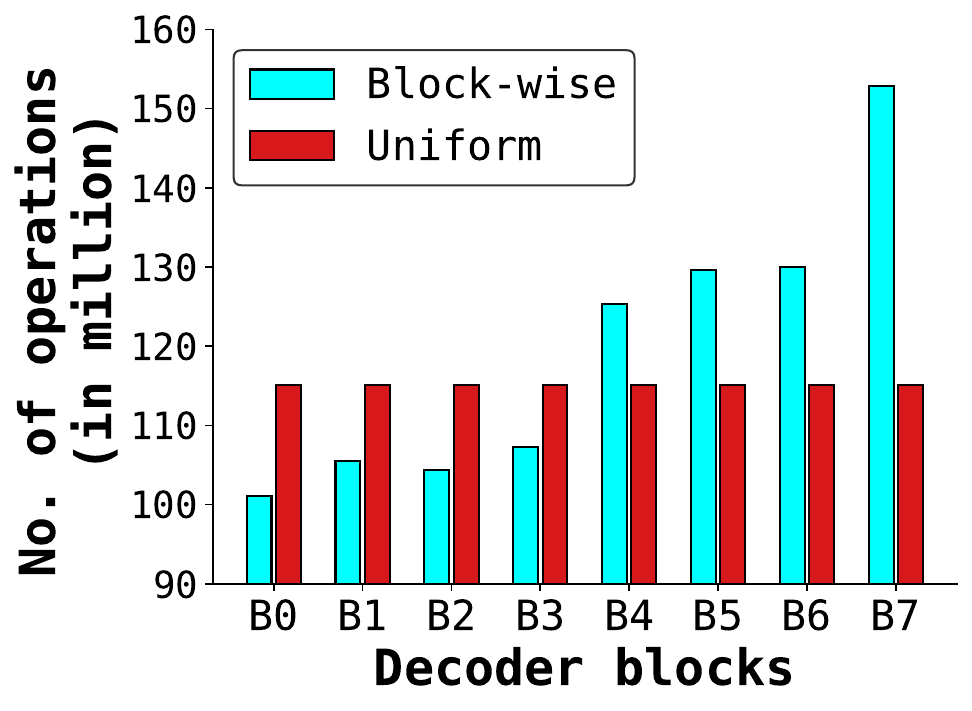}
        \end{tabular}
        \caption{Distribution of parameters and operations within each block}
    \end{subfigure}
    \caption{\textbf{Block-wise scaling} efficiently allocates parameters and operations across blocks, leading to shallower and narrower \arch~blocks near the input and deeper and wider \arch~blocks near the output. In (b), \arch~networks with both uniform ($N$=$N_{min}$=$N_{max}$=8) and block-wise ($N_{min}$=4, $N_{max}$=8) scaling have about 16.7 M parameters and perform 3.5 B operations (computed for a sequence length of $n=30$), however, the \arch~network with block-wise scaling delivered 2 points better perplexity.} 
    \label{fig:fixed_vs_layer_wise}
\end{figure}

Standard methods for improving the performance of sequence models include increasing the model dimensions (width scaling), stacking more blocks (depth scaling), or both. However, such scaling is not very effective on small datasets. For example, when a Transformer-Base ($d_m=512$)  network is replaced with Transformer-Large ($d_m=1024$) on the WMT'16 En-Ro corpus, the number of parameters increases by approximately $4\times$ while the performance does not change appreciably (BLEU: 34.28 vs. 34.35). We hypothesize that this happens because scaling model width and depth allocates parameters uniformly across blocks, which may lead to learning redundant parameters. To create deep and wide networks, we extend model scaling to the block level (see Figure \ref{fig:fixed_vs_layer_wise}).

\noindent{\bf Scaling the DeLighT block:} The \arch~block learns deep and wide representations using the \dextra, whose depth and width are controlled by two configuration parameters: the number of GLT layers $N$ and the width multiplier $w_m$, respectively (Figure \ref{fig:com_tam_ram}). These configuration parameters allow us to increase the number of learnable parameters inside the \arch~block independently of the input $d_m$ and output $d_o$ dimensions. Such calibration is not possible with the standard transformer block because their expressiveness and capacity are a function of the input (input dimension = number of heads $\times$ head dimension). Here, we introduce block-wise scaling that creates a network with variably-sized \arch~blocks, allocating shallower and narrower \arch~blocks near the input and deeper and wider \arch~blocks near the output.

To do so, we introduce two network-wide configuration parameters: minimum $N_{min}$ and maximum $N_{max}$ number of GLTs in a \dextra. For the $b$-th \arch~block, we  compute the number of GLTs $N^b$ and the width multiplier $w_m^b$ in a \dextra~using linear scaling (Eq. \ref{eq:mw}). With this scaling, each \arch~block has a different depth and width (Figure \ref{fig:com_tam_ram}). 
\begin{equation}
       N^{b} = N_{min} + \frac{(N_{max} - N_{min})\ b}{\mathcal{B}-1}, \quad w_m^{b} =  w_m + \frac{(N_{max} - N_{min})\ b}{N_{min}(\mathcal{B}-1)}, \quad 0 \le b \le \mathcal{B}-1
    \label{eq:mw}
\end{equation}
Here, $\mathcal{B}$ denotes the number of \arch~blocks in the network. We add superscript $b$ to number of GLT layers $N$ and width multiplier $w_m$ to indicate that these parameters are for the $b$-th block. 

\noindent{\bf Network depth:} The depth of transformer block is fixed, i.e., 4. Therefore, previous works \citep{raffel2019exploring,brown2020language,wang2019learning} have associated the depth of transformer-based networks with the number of transformer blocks. In \arch, we present a different perspective to learn deeper representations, wherein each block is variably-sized. To compute the network depth, we use the standard definition across different domains, including computer vision (e.g., ResNet of \citealt{he2016deep}) and theoretical machine learning \citep{telgarsky2016benefits}. These works measures network depth as the number of sequential learnable layers (e.g., convolution, linear, or group linear). Similarly, the depth of \arch~and transformer networks with $\mathcal{B}$ blocks is $\sum_{b=0}^{\mathcal{B}-1} (N^b + 4)$ and $4\mathcal{B}$, respectively.

\section{Experimental results}
\label{sec:results}
We evaluate the performance of \arch~on two standard sequence modeling tasks: (1) machine translation (Section \ref{ssec:machine_translation}) and (2) language modeling (Section \ref{ssec:language_mdoeling}).

\subsection{Machine Translation}
\label{ssec:machine_translation}
\noindent {\bf Datasets and evaluation:} We benchmark \arch~models on four datasets: (1) IWSLT'14 German-English (De-En), (2) WMT'16 English-Romanian (En-Ro), (3) WMT'14 English-German (WMT'14 En-De), and (4) WMT'14 English-French (WMT'14 En-Fr). For the IWSLT'14 De-En dataset, we replicate the setup of \citet{wu2018pay} and \citet{edunov2018classical}, which uses 160K/7K/7K sentence pairs for training, validation, and testing with a joint BPE vocabulary of about 10K tokens, respectively. For the WMT'14 English-German (En-De) dataset, we follow the setup of \citet{vaswani2017attention}. The dataset has 3.9M/39K/3K sentence pairs for training, validation, and testing respectively with a joint BPE vocabulary size of 44K.\footnote{We use training and validation data that is compatible with the Tensor2Tensor library \citep{tensor2tensor} in order to have fair comparisons with recent works (e.g., Evolved Transformer).} For the WMT'14 English-French (En-Fr) dataset, we replicate the setup of \citet{gehring2017convolutional}, which uses 36M/27K/3K sentence pairs for training, validation, and testing respectively with a joint BPE vocabulary size of 44K. The performance is evaluated in terms of \textit{BLEU} \citep{papineni2002bleu} (higher is better) on the test set. We follow \citet{wu2018pay} for beam search related hyper-parameters.

\noindent {\bf Architecture:} We follow the symmetric encoder-decoder architecture of \citet{vaswani2017attention} with sinusoidal positional encodings. Both the encoder and the decoder have $\mathcal{B}$ \arch~blocks. Decoder blocks are identical to the encoder blocks (Figure \ref{fig:redefine_transformer_sa}), except that they have an additional source-target single-head attention unit before the light-weight FFN. In the source-target single-head attention unit, keys and values are projections over the encoder output (full details in Appendix \ref{sec:appendix_enc_dec_arch}).  In our experiments, we use $w_m=2$, $N_{min}=4$, and $N_{max}=8$ for WMT'16 En-Ro, WMT'14 En-De, and WMT'14 En-Fr; resulting in 222 layer deep \arch~networks. For IWSLT'14 De-En, we used $w_m=1$, $N_{min}=3$, and $N_{max}=9$ for IWSLT'14 De-En; resulting in 289 layer deep network. For simplicity, we set $\mathcal{B}=N_{max}$. We use a learnable look-up table that maps every token in the vocabulary to a 128-dimensional vector. We implement our models using Fairseq \citep{ott2019fairseq} and use their provided scripts for data pre-processing, training, and evaluation.

\noindent {\bf Training:} For IWSLT'14 De-En models, we follow the setup of \cite{wu2018pay} and train all our models for 50K iterations with a batch size of 4K tokens on a single NVIDIA GTX 1080 GPU. For WMT'16 En-Ro, we follow the training setup of \cite{ghazvininejad2019mask} and train models for 100K iterations on 16 NVIDIA Tesla V100 GPUs with an effective batch size of 64K tokens. For WMT'14 En-De and WMT'14 En-Fr, we follow the training set-up of \cite{wu2018pay} and train our models on 16 V100 GPUs for 30K and 50K iterations, respectively. We use Adam \citep{kingma2014adam} to minimize cross entropy loss with a label smoothing value of 0.1 during training. For a fair comparison, we trained baseline transformer models using the same training set-up.

\subsubsection{Results}

\begin{figure}[t!]
    \begin{minipage}[b]{0.46\columnwidth}
        \vspace{0pt}
        \centering
        \resizebox{\columnwidth}{!}{
        \begin{tabular}{lrrrr}
            \toprule[1.5pt]
             & \textbf{Depth} & \textbf{\# Params} & \textbf{\# MACs} & \textbf{BLEU}  \\
             \midrule[1.25pt]
            Transformer & 60 & 67 M & 11.1 B & 39.2\\
            \arch & 222 & 37 M  & 5.6 B & 39.6 \\
            \arch & 222 & 54 M & 8.1 B & 40.5 \\
            \bottomrule[1.5pt]
        \end{tabular}
    }
        \captionof{table}{\textbf{\arch~networks are deep, light-weight and efficient} as compared to transformers. BLEU score is reported on the WMT'14 En-Fr dataset. To compute network depth, we count the number of sequential layers in the network (Section \ref{ssec:layer_wise_scaling}). We used 20 source and 20 target tokens for computing multiplication-addition operations (MACs). See Appendex \ref{sec:appendix_mac} for details.}
        \label{tab:results_depth_macs}
    \end{minipage}
    \hfill
    \begin{minipage}[b]{0.5\columnwidth}
        \vspace{0pt}
        \centering
        \includegraphics[height=85px]{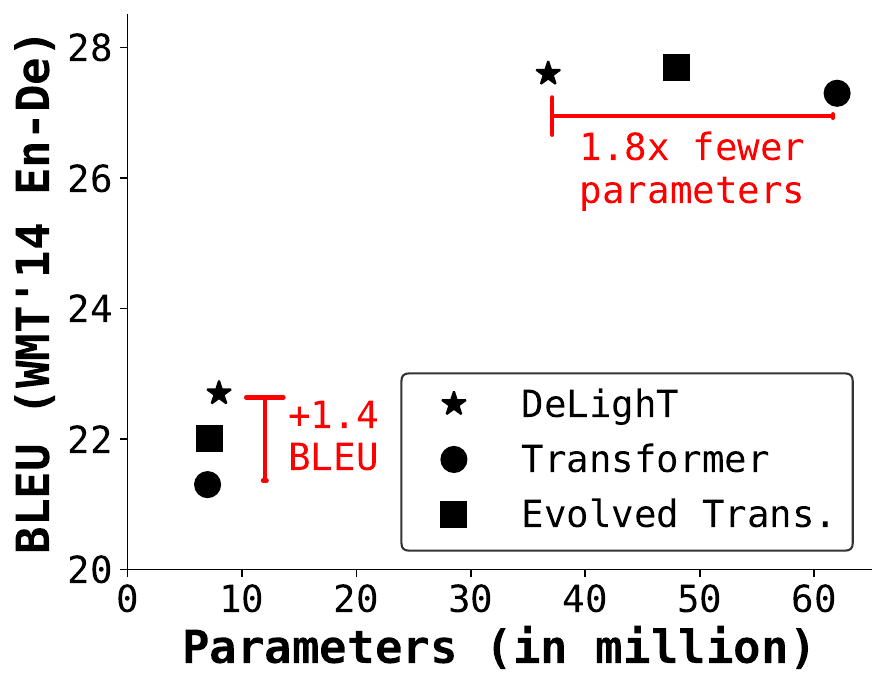}
        \caption{Comparison of \arch~with Transformers and Evolved Transformers at two different settings, on the WMT'14 En-De corpus: (1) the number of parameters is the same and (2) the performance is the same.}
        \label{fig:perf_compare_param}
    \end{minipage}
\end{figure}

\noindent {\bf Comparison with baseline transformers:} Table \ref{tab:compare_delight_trans} compares the performance of \arch~with the baseline transformers of \citet{vaswani2017attention} on different corpora. \arch~delivers better performance with fewer parameters than transformers, across different corpora. Specifically, on low-resource (WMT'16 En-Ro) and high resource (WMT'14 En-De \& WMT'14 En-Fr) corpora, \arch~delivers similar or better performance with $2.8\times$ and $1.8\times$ fewer parameters, respectively. When the number of parameters are increased, \arch~outperforms transformers. For example, on WMT'14 En-Fr dataset, \arch~is $3.7\times$ deeper than transformers and improves its BLEU score by 1.3 points yet with 13 million fewer parameters and 3 billion fewer operations (see Table \ref{tab:results_depth_macs}).

Particularly interesting are the performance comparisons of \arch~with the baseline transformers of \citet{vaswani2017attention} and its neural search variant, i.e., Evolved Transformer of \citet{so2019evolved}, at two different parametric settings on WMT'14 En-De corpora in Figure \ref{fig:perf_compare_param}. For small models (< 10 M parameters), \arch~models delivers better performance and for attaining the same performance as these models, \arch~models requires fewer parameters. 

\begin{table}[t!]
    \centering
    \begin{subtable}[b]{\columnwidth}
        \centering
    \resizebox{0.88\columnwidth}{!}{
        \begin{tabular}{lcc||cc c cc||cc}
            \toprule[1.5pt]
            \multicolumn{1}{c}{} &  \multicolumn{4}{c}{\textbf{IWSLT'14 De-En}} & \hfill & \multicolumn{4}{c}{\textbf{WMT'16 En-Ro}} \\
            \cmidrule[1.25pt]{2-5} \cmidrule[1.25pt]{7-10}
            \textbf{Model} & \textbf{\# Params} & \textbf{Ratio} & \textbf{BLEU} & $\Delta$ \textbf{BLEU} &\hfill & \textbf{\# Params} & \textbf{Ratio} & \textbf{BLEU} & $\Delta$ \textbf{BLEU} \\
            \midrule
            Transformer \citep{vaswani2017attention} & -- & -- & 34.4$^\dagger$ & -- && 62 M & -- & 34.3$^\ddagger$ & -- \\
            \midrule
            Transformer (Our impl.) & 42 M & $1.0\times$ & 34.3 & -- && 62 M & $1.0\times$ & 34.3 & -- \\
            \arch                   & 14 M & $0.3\times$ & 33.8 & -0.5 && 22 M & $0.35\times$ & 34.3 & 0.0 \\
            \arch                   & 30 M & $0.7\times$ & \textbf{35.3} & \textbf{+1.0} && 53 M & $0.85\times$ & \textbf{34.7} & \textbf{+0.4} \\
            \bottomrule
        \end{tabular}
    }
    \caption{Results on small corpora}
    \end{subtable}
    \vfill
    \begin{subtable}[b]{\columnwidth}
        \centering
    \resizebox{0.88\columnwidth}{!}{
        \begin{tabular}{lcc||cc p{0.25cm} cc||cc}
            \toprule[1.5pt]
            \multicolumn{1}{c}{} &  \multicolumn{4}{c}{\textbf{WMT'14 En-De}} & \hfill & \multicolumn{4}{c}{\textbf{WMT'14 En-Fr}} \\
            \cmidrule[1.25pt]{2-5} \cmidrule[1.25pt]{7-10}
            \textbf{Model} & \textbf{\# Params} & \textbf{Ratio} & \textbf{BLEU} & $\Delta$ \textbf{BLEU} &\hfill & \textbf{\# Params} & \textbf{Ratio} & \textbf{BLEU} & $\Delta$ \textbf{BLEU} \\
            \midrule
            Transformer \citep{vaswani2017attention} & 62 M & -- & 27.3 & -- && -- & 62 M & 38.1 & -- \\
            \midrule
            Transformer (Our impl.) & 67 M & $1.0\times$ & 27.7 & -- && 67 M & $1.0\times$ & 39.2 & -- \\
            \arch                   & 37 M & $0.55\times$ & 27.6 & -0.1 && 37 M & $0.55\times$ & 39.6 & +0.4 \\
            \arch                   & 54 M & $0.80\times$ & \textbf{28.0} & \textbf{+0.3} && 54 M & $0.80\times$ & \textbf{40.5} & \textbf{+1.3} \\
            \bottomrule[1.5pt]
        \end{tabular}
    }
    \caption{Results on large corpora}
    \end{subtable}
    \caption{\textbf{Comparison with baseline transformers on machine translation corpora}. \arch~models require significantly fewer parameters to achieve similar performance. Here, $^\dagger$ and $^\ddagger$ indicate the best reported transformer baselines from \citet{wu2018pay} and \citet{ghazvininejad2019mask}, respectively.}
    \label{tab:compare_delight_trans}
\end{table}

\noindent {\bf Comparison with \sota~methods:} Most \sota~methods have evaluated the performance on WMT'14 En-De while some have also evaluated on IWSLT'14 De-En. Table \ref{tab:comapre_nmt_sota} compares the performance of \arch~with \sota~methods on these two corpora. \arch~delivers similar or better performance than existing methods. It is important to note that existing methods have improved baseline transformers with different design choices -- for example, the asymmetric encoder-decoder structure \citep{wang2019learning} and neural architecture search \citep{so2019evolved}. We believe that \arch, in the future, would also benefit from such design choices. 

\begin{table}[t!]
    \centering
    \begin{subtable}[b]{0.52\columnwidth}
        \resizebox{0.9\columnwidth}{!}{
        \begin{tabular}{lrr}
            \toprule[1.5pt]
          \textbf{Model} & \textbf{\# Params} & \textbf{BLEU} \\
          \midrule
            Transformers \citep{vaswani2017attention} & 42 M & 34.3 \\
            Variational Attention \citep{deng2018latent} & -- & 33.1 \\
             Dynamic convolutions \citep{vaswani2017attention} & 43 M & \textbf{35.2} \\
             Lite Transformer$^\ddagger$ \citep{Wu2020Lite} & -- & 33.6 \\
             \arch~(Ours) & \textbf{30 M} & \textbf{35.3} \\
            \bottomrule[1.5pt]
        \end{tabular}
        }
        \caption{IWSLT'14 De-En}
    \end{subtable}
    \hfill
    \begin{subtable}[b]{0.47\columnwidth}
        \resizebox{0.9\columnwidth}{!}{
        \begin{tabular}{lrr}
            \toprule[1.5pt]
          \textbf{Model} & \textbf{\# Params} & \textbf{BLEU} \\
          \midrule
            Transformer \citep{vaswani2017attention} & 62 M & 27.3 \\
            DLCL \citep{wang2019learning} & 62 M & 27.3 \\
            Evolved Transformer $^\dagger$ \citep{so2019evolved} & 46 M & \textbf{27.7} \\
            Lite Transformer$^\ddagger$ \citep{Wu2020Lite} & -- & 26.5 \\
            \arch~(Ours) & \textbf{37 M} & \textbf{27.6} \\
            \bottomrule[1.5pt]
        \end{tabular}
        }
        \caption{WMT'14 En-De}
    \end{subtable}
    \caption{\textbf{Comparison with \sota~methods on machine translation corpora}. \arch~delivers similar or better performance than \sota~models with fewer parameters. Here, $\dagger$ indicates that the network uses neural architecture search (NAS) and $\ddagger$ indicates that full network parameters are not reported. }
    \label{tab:comapre_nmt_sota}
\end{table}

\begin{figure}[t!]
    \centering
    \begin{subfigure}[b]{0.24\columnwidth}
        \centering
        \includegraphics[width=\columnwidth]{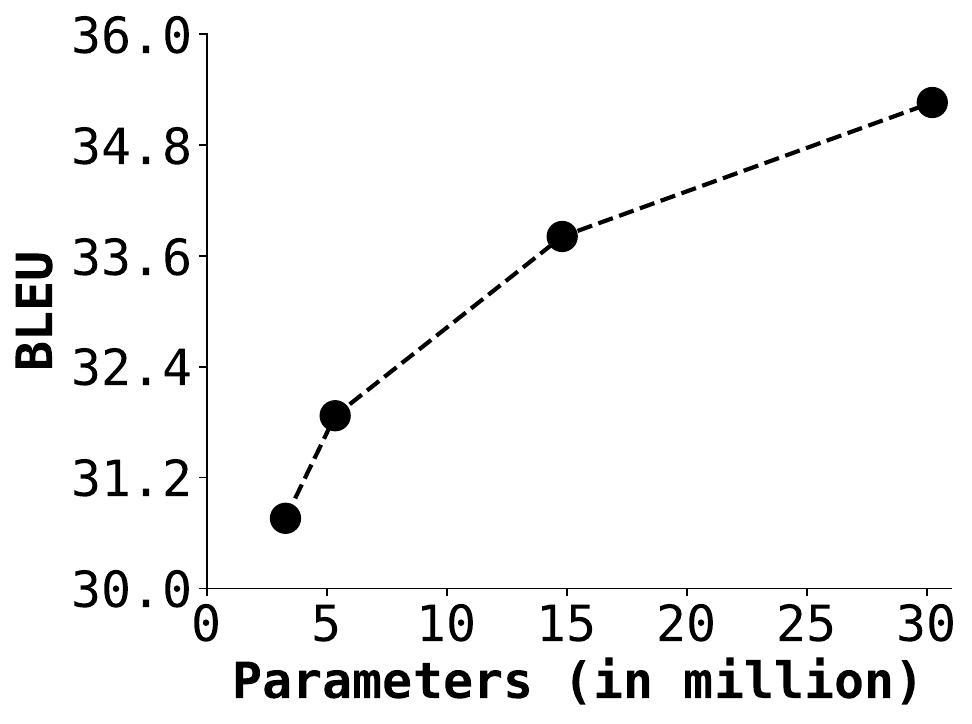}
        \caption{IWSLT'14 De-En}
    \end{subfigure}
    \hfill
    \begin{subfigure}[b]{0.24\columnwidth}
        \centering
        \includegraphics[width=\columnwidth]{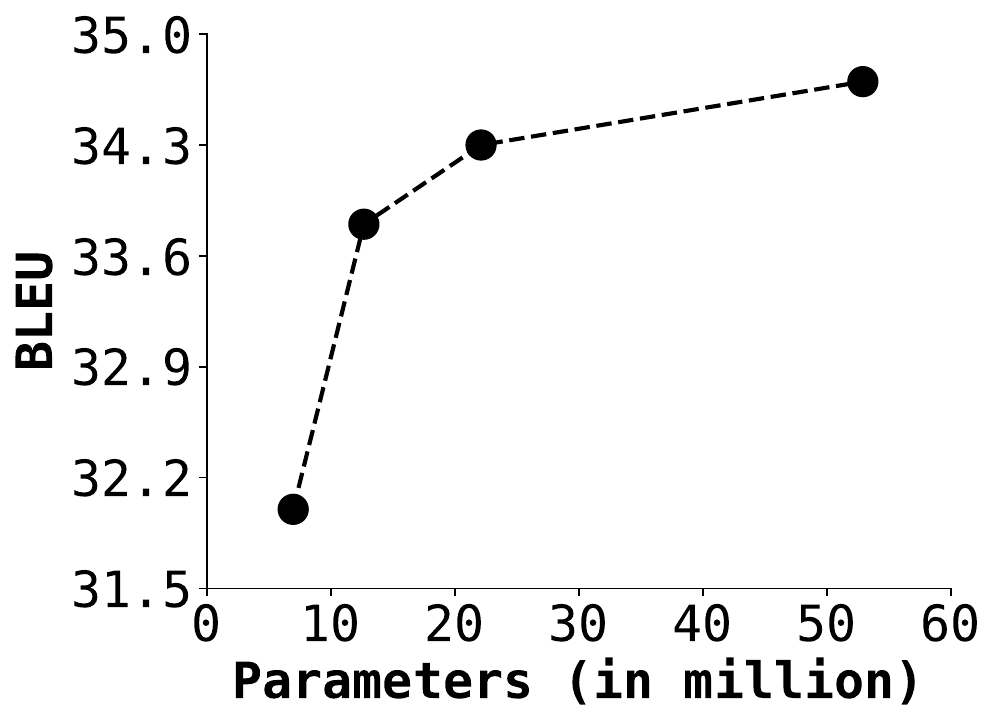}
        \caption{WMT'16 En-Ro}
    \end{subfigure}
    \hfill
    \begin{subfigure}[b]{0.24\columnwidth}
        \centering
        \includegraphics[width=\columnwidth]{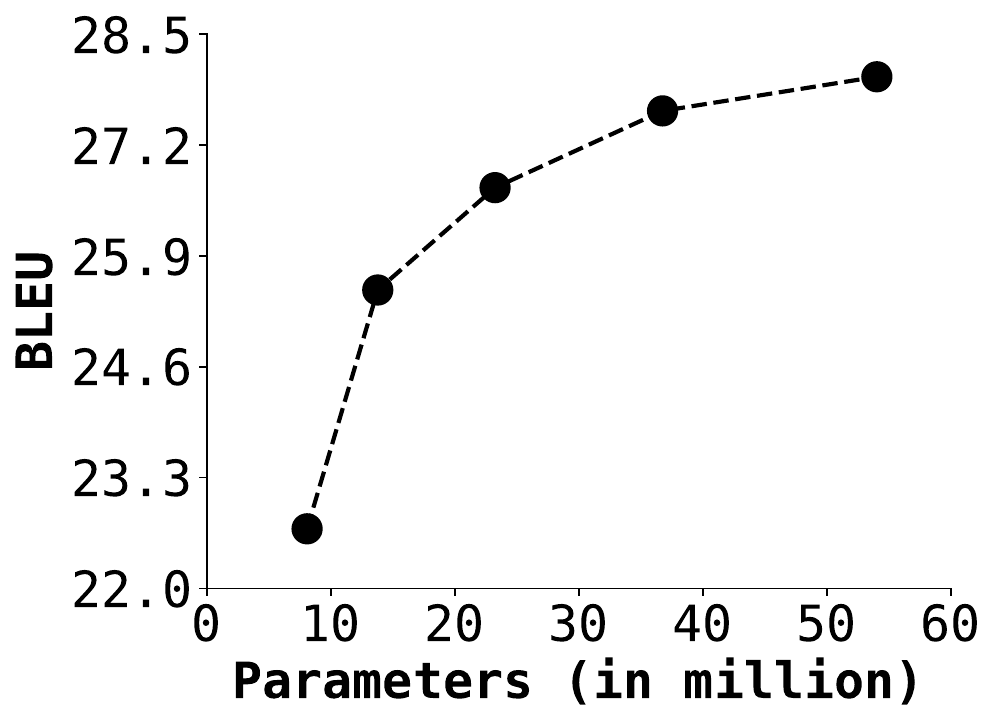}
        \caption{WMT'14 En-De}
    \end{subfigure}
    \hfill
    \begin{subfigure}[b]{0.24\columnwidth}
        \centering
        \includegraphics[width=\columnwidth]{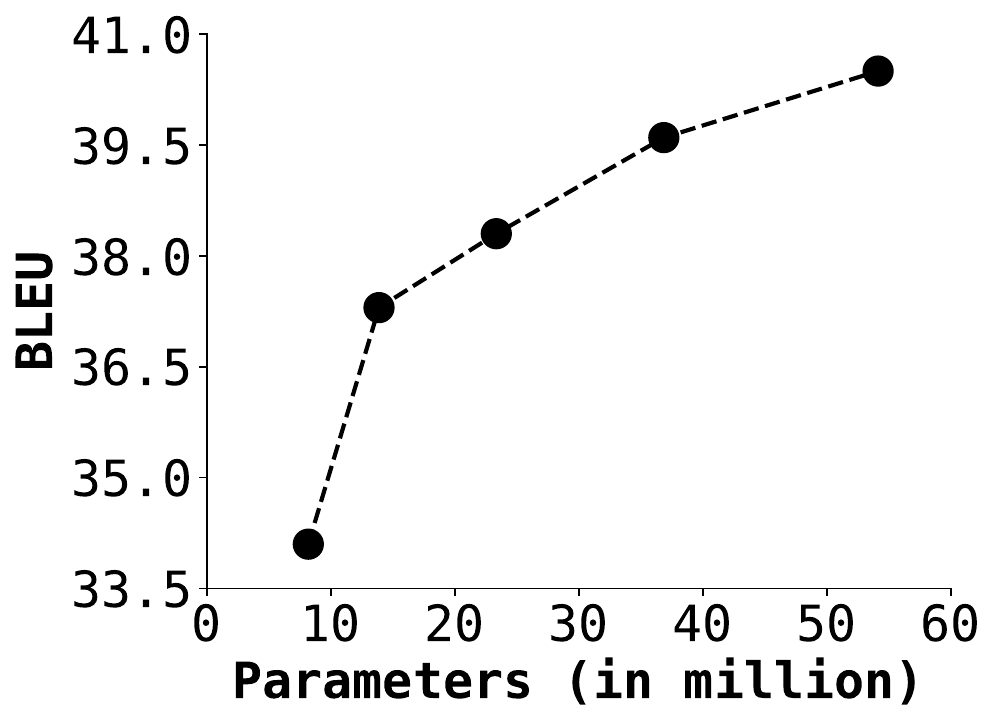}
        \caption{WMT'14 En-Fr}
    \end{subfigure}
    \caption{\textbf{Scaling up \arch~models.} The performance of \arch~improves with an increase in the number of network parameters, across different corpora, including low-resource (WMT'16 En-Ro).}
    \label{fig:nmt_perf_curve}
\end{figure}

\noindent {\bf Scaling up \arch~models:} Figure \ref{fig:nmt_perf_curve} shows the performance of \arch~models improves with increase in network parameters; suggesting their ability to learn representations across different corpora, including low-resource.

\subsection{Language Modeling}
\label{ssec:language_mdoeling}

\noindent {\bf Datasets and evaluation:} We evaluate on the WikiText-103 dataset \citep{merity2017pointer} that has 103M/217K/245K tokens for training, validation, and testing. It has a word-level vocabulary of about 260K tokens. Following recent works \citep{baevski2018adaptive,dai2019transformer}, we report performance in terms of \textit{perplexity} (lower is better) on the test set.

\noindent {\bf Architecture:} We use the transformer-based decoder architecture of \citet{baevski2018adaptive} with $\mathcal{B}$ \arch~blocks. We use $w_m$=$2$, $N_{min}$=$4$, and $N_{max}$=$12$. We scale $d_m$ using values $\{384, 512, 784, 1024\}$ for increasing network parameters. For simplicity, we set $\mathcal{B}=N_{max}$. Following standard practice, we use adaptive input \citep{baevski2018adaptive} as a look-up table and adaptive output \citep{grave2017efficient} as the classification layer with one head (head dimension is 128) and two tails (tail dimensions are 64 and 32). We also share weights between the input and the output layers.

\vspace{0.5mm}
\noindent {\bf Training:} We follow the training setup of \citet{baevski2018adaptive}, except that we train our models on 8 NVIDIA Tesla V100 GPUs for 100K iterations with a context length of 512 and an effective batch size of 64K tokens. We use Adam during training and use a context length of 480 during test.

\noindent {\bf Results:} Table \ref{tab:wiki_sota} compares the performance of \arch~with previous methods on WikiText-103. Table \ref{tab:wiki_txl} plots the variation of perplexity with number of parameters for \arch~and Transformer-XL \citep{dai2019transformer} -- which outperforms other transformer-based implementations (e.g., \citealt{baevski2018adaptive}). Both tables show that \arch~delivers better performance than \sota~methods (including Transformer-XL) and it does this using a smaller context length and significantly fewer parameters, suggesting that the \dextra~helps learn strong contextual relationships.

%\noindent {\bf Ablations:} We also did an extensive set of ablation studies to better understand which parts of the architecture contribute the most to improve performance and we include these in the Appendix~\ref{sec:appendix_ablations}. We varied different hyper-parameters including minimum $N_{min}$ and maximum $N_{max}$ number of GLTs, width multiplier $w_m$, and model dimension $d_m$. Overall, the differing settings either hurt performance or increase the parameter count with no further performance gains. 

\begin{table}[t!]
    \centering
    \begin{subtable}[b]{0.33\columnwidth}
        \centering
        \includegraphics[width=0.8\columnwidth]{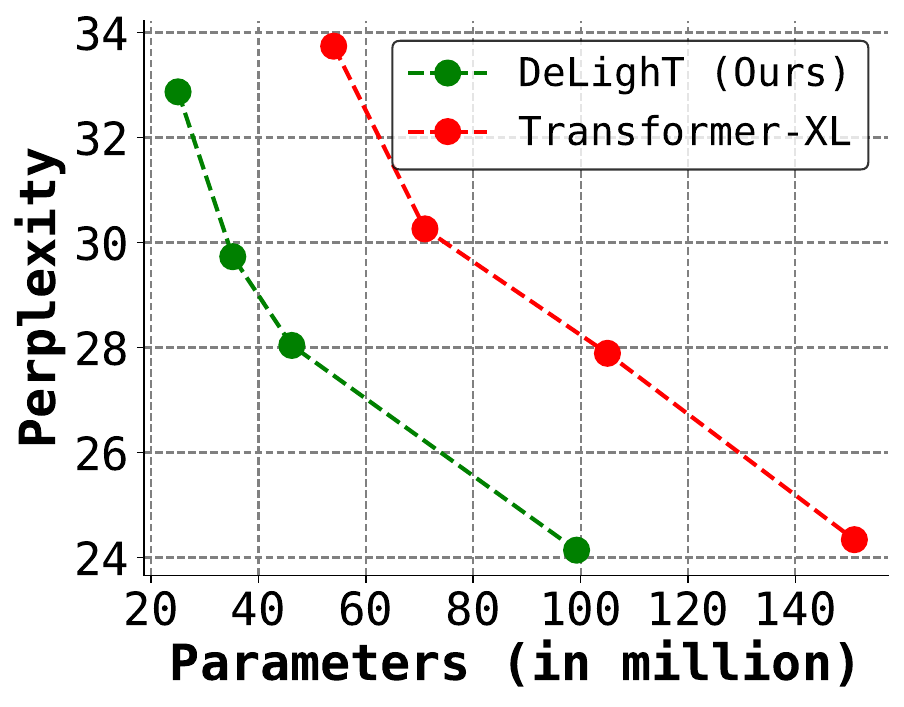}
        \caption{\arch~vs. Transformer-XL}
        \label{tab:wiki_txl}
    \end{subtable}
    \hfill
    \begin{subtable}[b]{0.65\columnwidth}
        \centering
        \resizebox{\columnwidth}{!}{
        \begin{tabular}{lcccc}
            \toprule[1.5pt]
           \multirow{2}{*}{\textbf{Method}} & \textbf{Network} & \textbf{Context}  & \textbf{\# Params} & \textbf{Perplexity} \\
            & \textbf{Depth} & \textbf{Length}  & \textbf{ (in million)} & \textbf{(Test)} \\
           \midrule[1pt]
            LSTM \citep{grave2016improving} &--& -- & -- & 48.70 \\
            LSTM + Neural Cache \citep{grave2016improving} &--&-- & --& 40.80 \\
            QRNN \citep{merity2018analysis} &--& -- & 151 M & 33.00 \\
            \midrule
            Transformer-XL \citep{dai2019transformer} & 64 & 640 & 151 M & \textbf{24.03} \\
            Transformer-XL (Our impl.)$^\dagger$  & 64 & 640 & 151 M & 24.34 \\
            Transformer-XL (Our impl.)$^\dagger$  & 64 & 480 & 151 M & 24.91 \\
            \arch~(Ours) & 158 & \textbf{480} & \textbf{99 M} & \textbf{24.14} \\
            \bottomrule[1.5pt]
        \end{tabular}
        }
        \caption{Comparison with existing methods}
        \label{tab:wiki_sota}
    \end{subtable}
    \caption{\textbf{Results on the WikiText-103 dataset}. Compared to Transformer-XL, \arch~delivers similar or better performance (lower perplexity) with fewer parameters. $^\dagger$For Transformer-XL, we reproduce results using the official source code. For evaluating Transformer-XL with a context length of 480, we set the mem\_len hyper-parameter to 480 in the official evaluation scripts.}%\protect\footnotemark }}
    \label{tab:wiki103_results}
\end{table}

\begin{table}[t!]
    \centering
    \resizebox{0.75\columnwidth}{!}{
    \begin{tabular}{llcccc}
        \toprule[1.5pt]
        \multirow{2}{*}{\textbf{Row \#}} & \multirow{2}{*}{\textbf{Model}} & \textbf{\# Params} & \textbf{BLEU} & \textbf{Training} & \textbf{Memory} \\
         &  & \textbf{(in million)} & \textbf{(WMT'14 En-Fr)} & \textbf{time} & \textbf{(in GB)} \\
        \midrule[1pt]
        R1 & Transformer (unoptimized) & 67 M & 39.2 & 37 hours & 12.5 GB \\
        R2 & \arch~(unoptimized) & 54 M & 40.5 & 23 hours & 14.5 GB \\
        R3 & Transformer (w/ Apex optimized) & 67 M & 39.2 & 16 hours & 11.9 GB \\
        R4 & \arch~(w/ optimized grouping) & 54 M & 40.5 & 19 hours & 11.5 GB \\
        \bottomrule[1.5pt]
    \end{tabular}
    }
    \caption{Comparison with baseline transformers in terms of training speed and memory consumption. In R4, we implemented CUDA kernels for grouping and ungrouping functions only (see Appendix \ref{sec:append_source}). We expect \arch~to be more efficient with a single and dedicated CUDA kernel for grouping, transformation, feature shuffling, and ungrouping. Memory consumption is measured on a single NVIDIA GP100 GPU (16 GB memory) with a maximum of 4096 tokens per batch and without any gradient accumulation.}
    \label{tab:compute_compare}
\end{table}

\begin{table}[t!]
    \centering
    \resizebox{0.35\columnwidth}{!}{
        \begin{tabular}{lcc}
            \toprule[1.5pt]
           \textbf{Model}  & \textbf{Dropout} & \textbf{BLEU} \\
           \midrule[1pt]
            Transformer (62 M) & 0.10 & 27.3 \\
            Transformer (62 M) & 0.30 & 27.7 \\
            \arch~(37 M) & 0.05 & 27.6 \\
            \toprule[1.5pt]
        \end{tabular}
    }
    \caption{\arch~requires less regularization as compared to baseline transformers (Dataset: WMT'14 En-De).}
    \label{tab:dropout_effect}
\end{table}

\section{Analysis and Discussions on Computational Efficiency}

\noindent \textbf{Training time and memory consumption:} Table \ref{tab:compute_compare} compares the training time and memory consumption of \arch~with baseline transformers. For an apples-to-apples comparisons, we implemented the Transformer unit without NVIDIA's dedicated CUDA kernel, and trained both transformer and \arch~full-precision networks for 30K iterations on 16 NVIDIA V100 GPUs. The transformer and \arch~models took about 37 and 23 hours for training and consumed about 12.5 GB and 14.5 GB of GPU memory, respectively (R1 vs. R2). When we enabled the dedicated CUDA kernel provided by APEX library\footnote{https://github.com/NVIDIA/apex} for multi-head attention in Transformers, the training time of the transformer model reduced from 37 to 16 hours while we did not observe any significant change in memory consumption. Motivated by this observation, we implemented dedicated CUDA kernels for grouping and ungrouping functions in GLTs (see Appendix \ref{sec:append_source}). With these changes, training time and GPU memory consumption of \arch~reduced by about 4 hours and 3 GB, respectively. We emphasize that grouping, linear transformation, feature shuffling, and ungrouping, can be implemented efficiently using a single CUDA kernel. In future, we expect a dedicated CUDA kernel for these operations would further reduce the memory consumption as well as training/inference time.

\vspace{0.5mm}
\noindent \textbf{Regularization:} Table \ref{tab:dropout_effect} shows that \arch~delivers similar performance to  baseline transformers, but with fewer parameters and less regularization. This suggests that learning representations with better transformation functions alleviates the need for dropout.

\section{Conclusion}
This paper introduces a deep and light-weight transformer architecture, \arch, that efficiently allocates parameters both within the \arch~block and across \arch~blocks. Compared to state-of-the-art transformer models, \arch~models are (1) deep and light-weight and (2) deliver similar or better performance. In the future, we plan to apply \arch~to other tasks, including language model pre-training, question answering, and language generation.

\vspace{0.5mm}
\noindent \textbf{Acknowledgements:} This research was supported by ONR N00014-18-1-2826, DARPA N66001-19-2-403, NSF (IIS-1616112, IIS1252835), and an Allen Distinguished Investigator Award. Authors would also like to thank members of the UW-NLP and the H2Lab at The University of Washington for their valuable feedback and comments.

\small{
\bibliographystyle{unsrtnat}
\bibliography{main}
}

\clearpage

\appendix

\section{\arch~Architectures for Language Modeling and Machine Translation}
\label{sec:appendix_enc_dec_arch}
\arch~architectures for language modeling and machine translation are shown in Figure \ref{fig:appendix_arch}. For language modeling, we follow the architecture in \citet{baevski2018adaptive} while for machine translation, we follow the architecture in \citet{vaswani2017attention}. 

\begin{figure}[b!]
    \centering
    \begin{subfigure}[b]{0.38\columnwidth}
        \centering
        \resizebox{\columnwidth}{!}{
            \input{tikz/enc_dec.tikz}\decattn
        }
        \caption{Language Modeling}
        \label{fig:appendix_arch_lm}
    \end{subfigure}
    \hfill
    \begin{subfigure}[b]{0.6\columnwidth}
         \centering
        \resizebox{\columnwidth}{!}{
            \input{tikz/enc_dec.tikz}\encdecattn
        }
        \caption{Machine translation}
        \label{fig:appendix_arch_enc_dec}
    \end{subfigure}
    \caption{Sequence modeling with \arch. Here, \colorbox{green}{green color hexagon} represents the \dextra.}
    \label{fig:appendix_arch}
\end{figure}

\vspace{1mm}
\noindent {\bf Language modeling:} Figure \ref{fig:appendix_arch_lm} shows the architecture for language modeling. The architecture stacks $\mathcal{B}$ \arch~blocks, the configuration of each block is determined using block-wise scaling. Each block has three sub-layers. The first layer is a \dextra~that learns representations in high-dimensional space. The second layer is a single-head attention that encodes contextual relationships. The third layer is a position-wise light-weight feed-forward network. Similar to  \citet{vaswani2017attention}, we employ a residual connections \citep{he2016deep}. Similar to previous works \citep{baevski2018adaptive,dai2019transformer}, we use tied adaptive input \citep{baevski2018adaptive} and adaptive softmax \citep{grave2017efficient} to map tokens to vectors and vectors to tokens, respectively.

\vspace{1mm}
\noindent {\bf Machine translation:} Figure \ref{fig:appendix_arch_enc_dec} shows the architecture for machine translation. The encoder stacks $\mathcal{B}$ \arch~blocks, the configuration of each block is determined using block-wise scaling. Similar to language modeling, each encoder block has three sub-layers. The first layer is a \dextra~that learns representations in high-dimensional space. The second layer is a single-head attention that encodes contextual relationships. The third layer is a position-wise light-weight feed-forward network. Similar to \citet{vaswani2017attention}, we employ a residual connections \citep{he2016deep}. We use learnable look-up table to map tokens to vectors. Similar to the encoder, the decoder also stacks  $\mathcal{B}$ blocks. Decoder blocks are identical to encoder blocks, except that they have an additional source-target single-head attention unit before the light-weight FFN. Keys and values in source-target single-head attention unit are projections over the encoder output. We use standard learnable look-up table to map tokens to vectors and linear classification layer to map vectors to tokens. 

\section{Group linear transformation with Input-mixer connection}
\label{sec:glt_explain}
Group linear transformation (GLT) $\mathcal{F}$ splits a $d_m$-dimensional input $\mathbf{X}$ into $g$ non-overlapping groups such that $\mathbf{X} = \text{Concat}(\mathbf{X}_1, \cdots, \mathbf{X}_g)$, where $\mathbf{X}_i$ is a $\frac{d_m}{g}$-dimensional vector. $\mathbf{X}_i$'s are then simultaneously transformed using $g$ linear transforms $\mathbf{W}_i \in \mathbf{R}^{\frac{d_m}{g} \times \frac{d_o}{g}}$ to produce $g$ outputs $\mathbf{Y}_i = \mathbf{X}_i \mathbf{W}_i$. $\mathbf{Y}_i$'s are then concatenated to produce the final $d_o$-dimensional output $\mathbf{Y} = \text{Concat}(\mathbf{Y}_1, \cdots, \mathbf{Y}_g)$. 

\begin{figure}[h!]
    \centering
    \begin{subfigure}[b]{0.3\columnwidth}
    \includegraphics[height=200px]{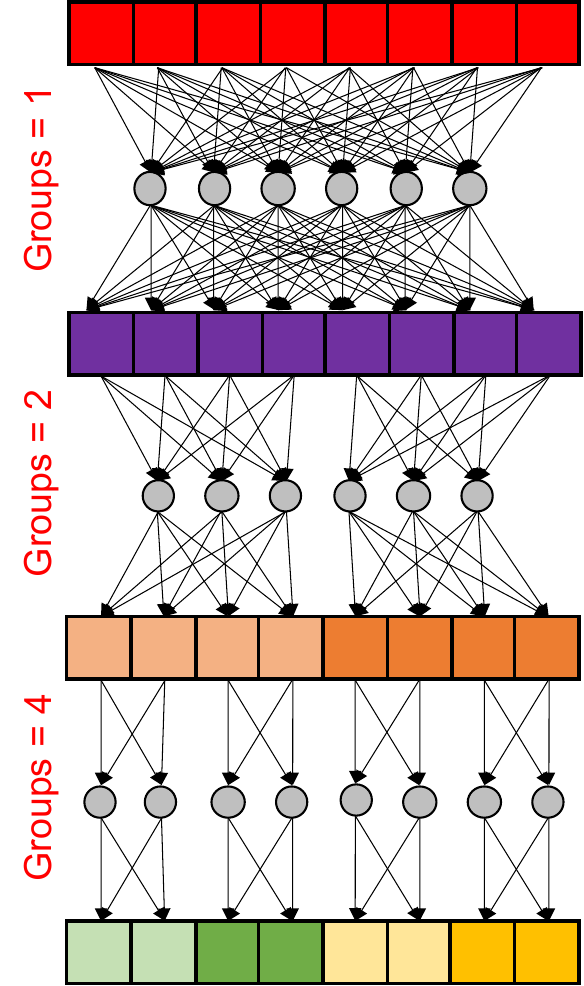}
    \caption{GLT}
    \label{fig:glt_app}
    \end{subfigure}
    \hfill
    \begin{subfigure}[b]{0.3\columnwidth}
        \includegraphics[height=200px]{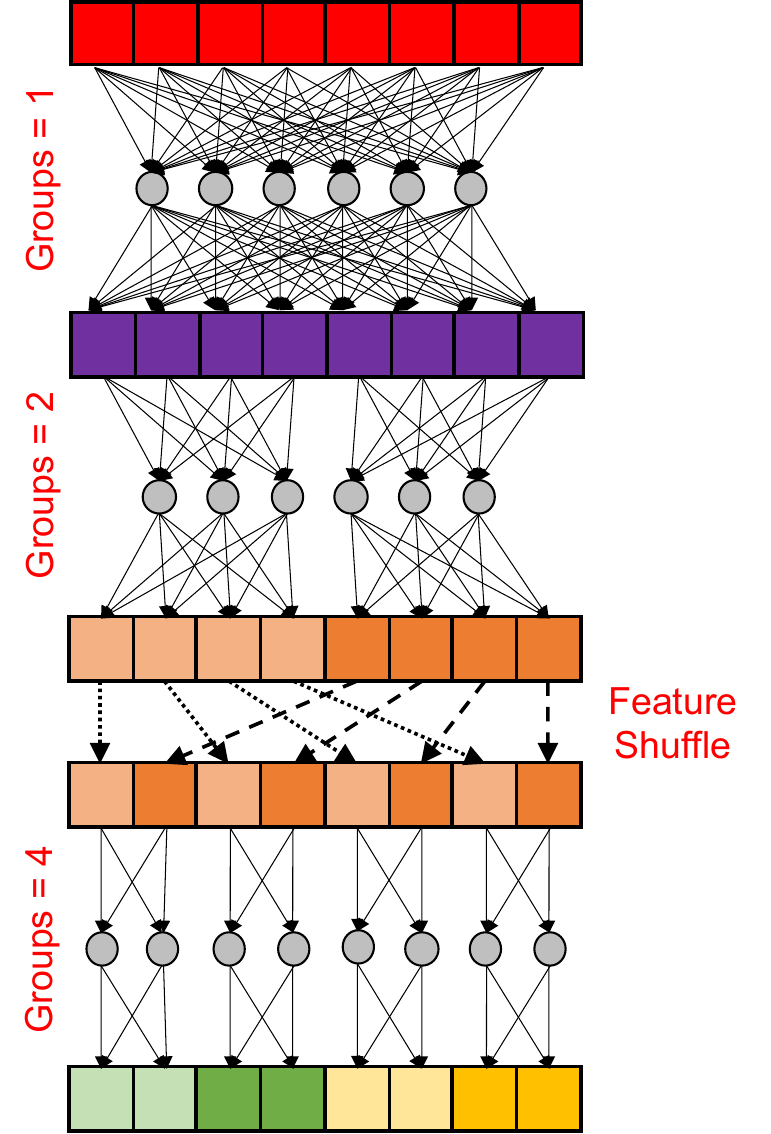}
        \caption{GLT with feature shuffle}
        \label{fig:glt_shuff_app}
    \end{subfigure}
    \hfill
    \begin{subfigure}[b]{0.33\columnwidth}
        \includegraphics[width=\columnwidth]{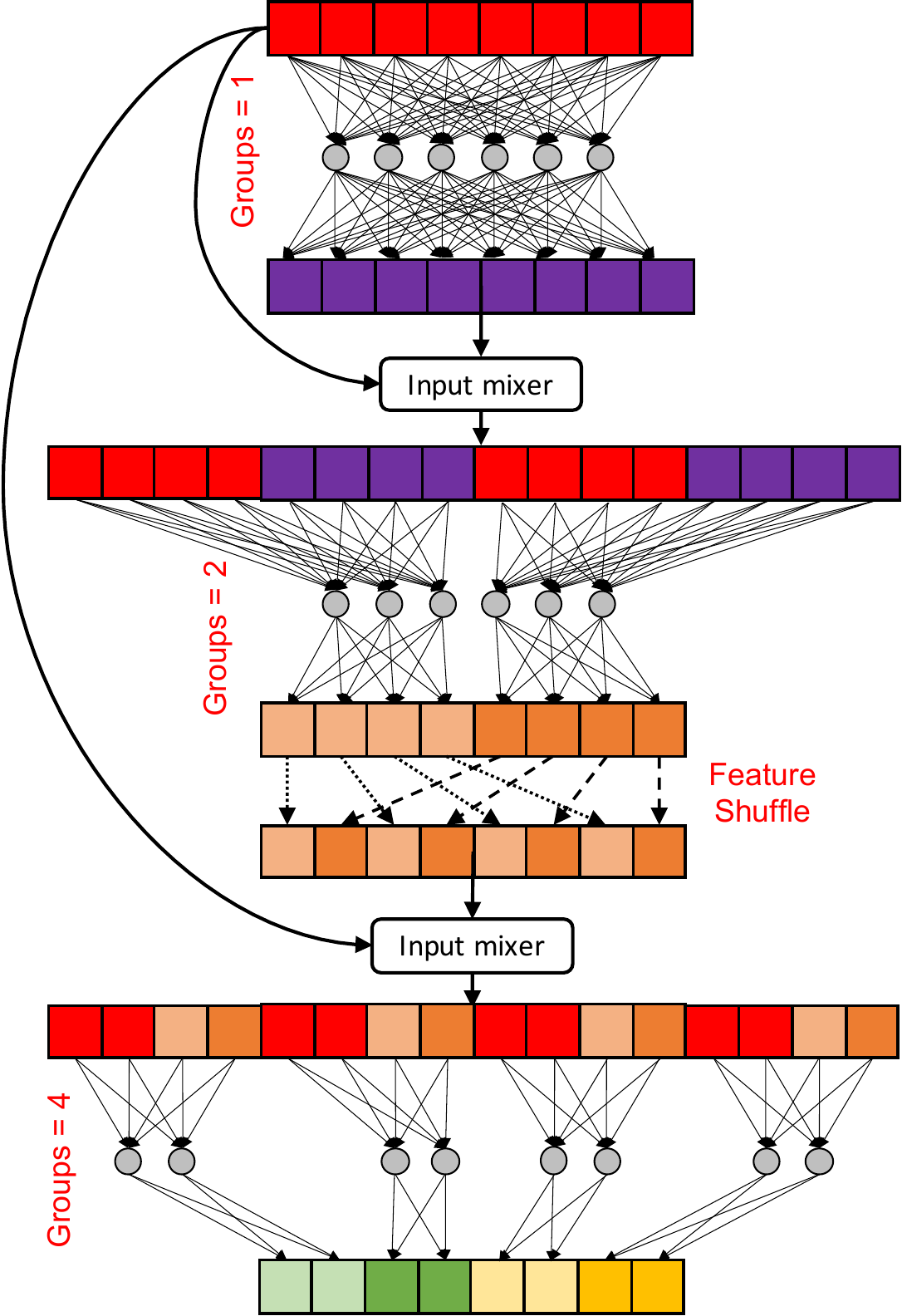}
        \caption{GLT with feature shuffle and input mixture connection}
        \label{fig:glt_shuff_mix_app}
    \end{subfigure}
    \caption{This figure visualizes different variants of group linear transformations that are used in the \dextra.}
    \label{fig:glt_vis}
\end{figure}

Figure \ref{fig:glt_app} shows an example of GLT in the expansion phase of \dextra. For illustrative purposes, we have used the same dimensions in this example. Recall that as we go deeper in the expansion phase, the number of groups increases. In this example, the first layer has one group, the second layer has two groups and the third layer has four groups. GLTs learns group-specific representations and are local.  To allow GLT to learn global representations, we use feature shuffle. An example of GLT with feature shuffle is shown in Figure \ref{fig:glt_shuff_app}. Furthermore, training deep neural networks by merely stacking linear or group linear (with or without feature shuffle) is challenging because of vanishing gradient problem. Residual connections introduced by \citet{he2016deep} mitigates this problem and helps train deep neural networks. However, such connections cannot be employed when input and output dimensions are not the same (e.g., during the expansion and reduction phases in \dextra). To stabilize the training and learn deeper representations, we use input-mixer connection of \citet{mehta2020DeFINE}. Figure \ref{fig:glt_shuff_mix_app} shows an example of GLT with feature shuffle and input mixer connection.

\section{Multiplication-Addition Operations in \arch}
\label{sec:appendix_mac}
The \arch~block is built using linear transformations, GLTs, and scaled dot-product attention. Total number of multiplication-addition operations (MACs) in a network is an accumulation of these individual operations. 

Let $n$ denotes the number of source tokens, $m$ denotes the number of target tokens, $d_m$ denotes the input dimension, $d_o$ denotes the output dimension, and $g$ denotes the number of groups in GLT. The procedure for counting MACs for each of these operations is described below. 

\paragraph{Group linear transformation (GLT):} GLT $\mathcal{F}$ has $g$ learnable matrices $\mathbf{W}_i \in \mathbf{R}^{\frac{d_m}{g} \times \frac{d_o}{g}}$. Therefore, GLT learns \colorbox{red!20}{$\frac{d_md_o}{g}$} parameters and performs \colorbox{red!20}{$\frac{d_md_o}{g}$} MACs to transform $d_m$-dimensional input to $d_o$-dimensional output. Following a standard practice, e.g., ResNet of \citet{he2016deep}, we count addition and multiplication as one operation instead of two because these operations can be fused in recent hardwares.

Importantly, when $g=1$, the GLT is the same as linear transformation.

\paragraph{Self-attention in \arch:} The scaled dot-product self-attention in \arch~is defined as:
\begin{equation}
    \text{Attention}(\mathbf{K}, \mathbf{Q}, \mathbf{V}) = \text{softmax}\left(\frac{\mathbf{Q}\mathbf{K}^T}{\sqrt{d_o}}\right) \mathbf{V}
\end{equation}
where $\mathbf{Q} \in \mathbb{R}^{n \times d_o}$, $\mathbf{K} \in \mathbb{R}^{n \times d_o}$, $\mathbf{V} \in \mathbb{R}^{n \times d_o}$ denotes query, key, and value, respectively.

The attention operation involves two dot-products. The first dot product between $\mathbf{Q}$ and $\mathbf{K}$ while the second dot product is between the output of first dot product and $\mathbf{V}$. Both dot products require $d_on^2$ MACs. Therefore, total number of MACs in computing scaled dot-product self-attention are \colorbox{red!20}{$2d_on^2$}.

In case of a source-target attention (as in machine translation), $\mathbf{K}$'s and $\mathbf{V}$'s are from the source (encoder) and $\mathbf{Q}$'s are incrementally decoded (one token at a time). Therefore, the number of MACs required to decode $m$ target tokens given $n$ source tokens are \colorbox{red!20}{$\sum\limits_{k=1}^{m} 2knd_o$}.

\section{Ablations on the WikiText-103 dataset}
\label{sec:appendix_ablations}
\begin{figure}[b!]
    \centering
    \begin{minipage}{0.45\columnwidth}
        \includegraphics[width=0.9\columnwidth]{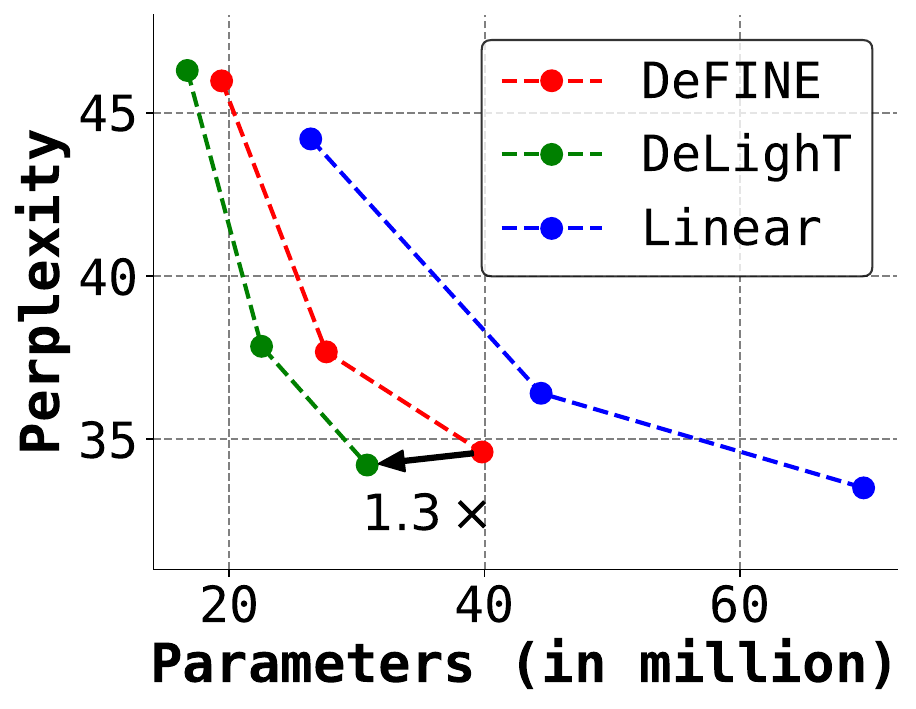}
        \caption{\textbf{Impact of different transformations.} \arch~transformations are more parametric efficient than DeFINE and linear transformations. Lower perplexity value means better performance.}
    \label{fig:dextra_define}
    \end{minipage}
    \hfill
    \begin{minipage}{0.45\columnwidth}
        \centering
        \includegraphics[width=0.9\columnwidth]{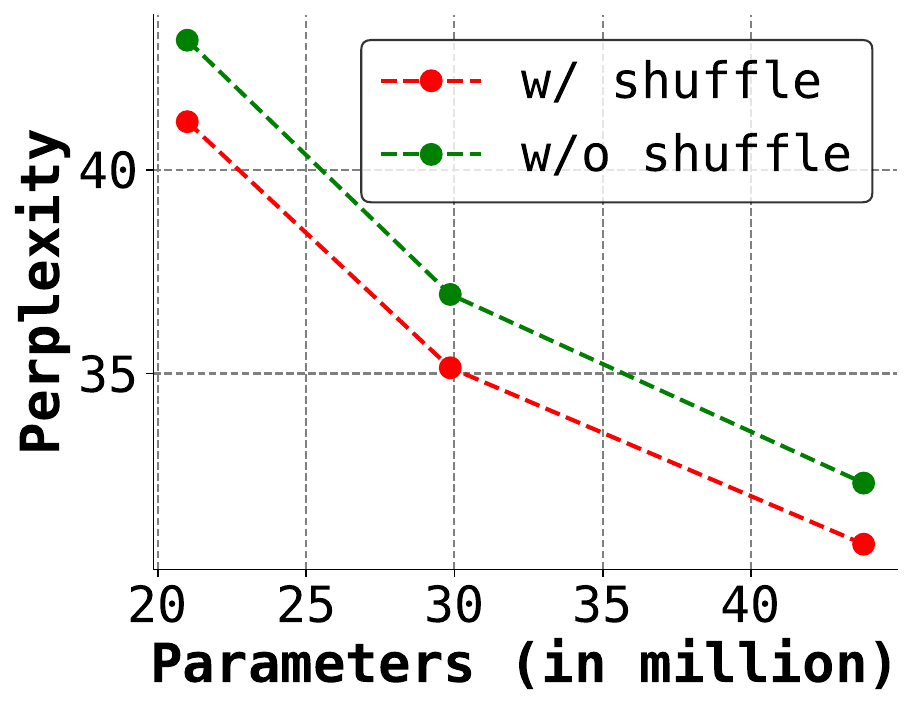}
        \caption{\textbf{Impact of feature shuffling.} Feature shuffling allows us to learn representations from global information and improves performance. Lower perplexity value means better performance.}
        \label{fig:shuffling}
    \end{minipage}
\end{figure}

\begin{figure}[t!]
    \centering
    \includegraphics[width=0.45\columnwidth]{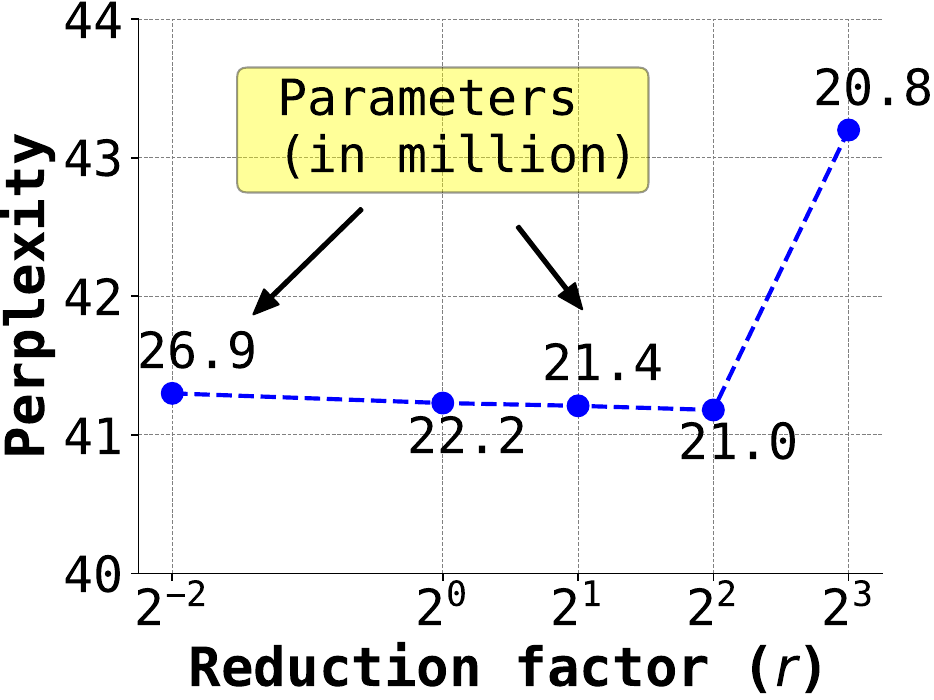}
    \caption{\textbf{Impact of reduction factor $r$ in light-weight FFN.} The ability of \dextra~to learn representations in high-dimensional spaces efficiently allows us to reduce the computational burden on the FFN. Lower perplexity value means better performance.}
    \label{fig:reduction_fac}
\end{figure}

\begin{figure}[t!]
    \centering
        \begin{subfigure}[b]{0.49\columnwidth}
            \centering
            \includegraphics[width=0.9\columnwidth]{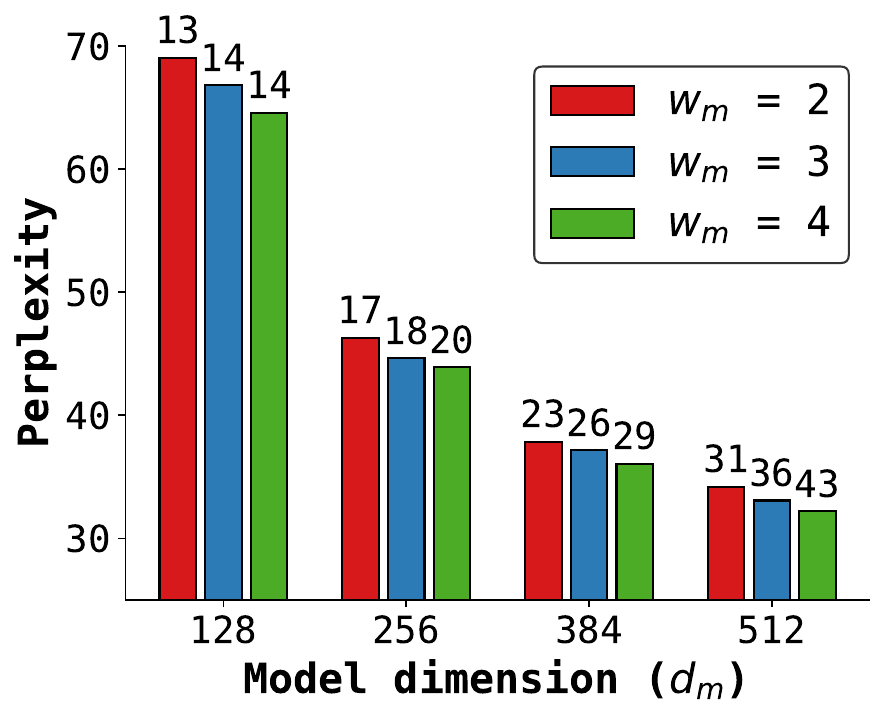}
            \caption{$N_{min}$=4, $N_{max}$=8}
            \label{fig:arch_4_8}
        \end{subfigure}
        \hfill
         \begin{subfigure}[b]{0.49\columnwidth}
            \centering
            \includegraphics[width=0.9\columnwidth]{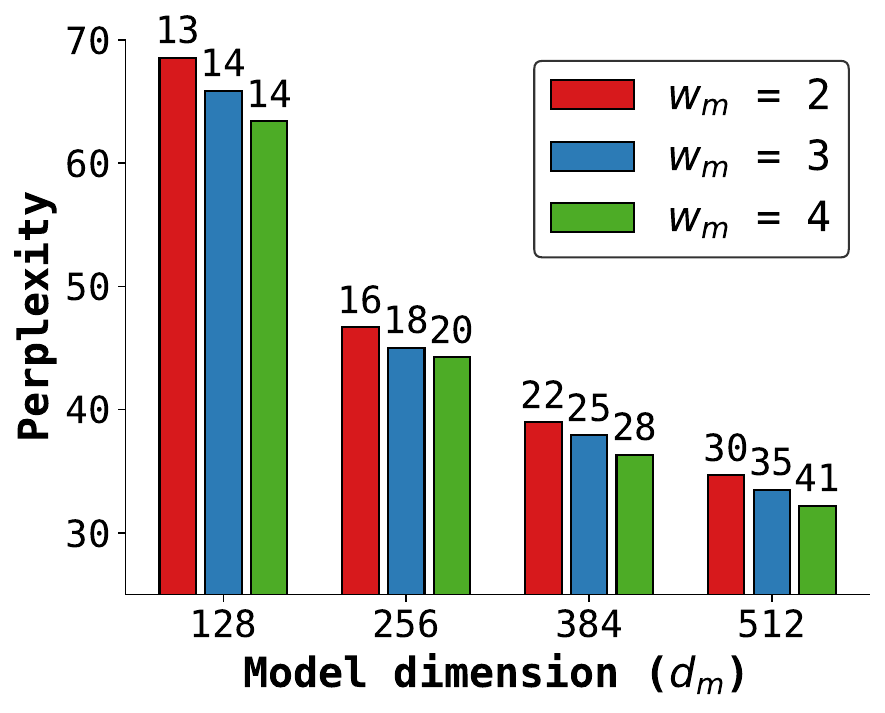}
            \caption{$N_{min}$=6, $N_{max}$=8}
            \label{fig:arch_6_8}
        \end{subfigure}
        \vfill
         \begin{subfigure}[b]{0.49\columnwidth}
            \centering
            \includegraphics[width=0.9\columnwidth]{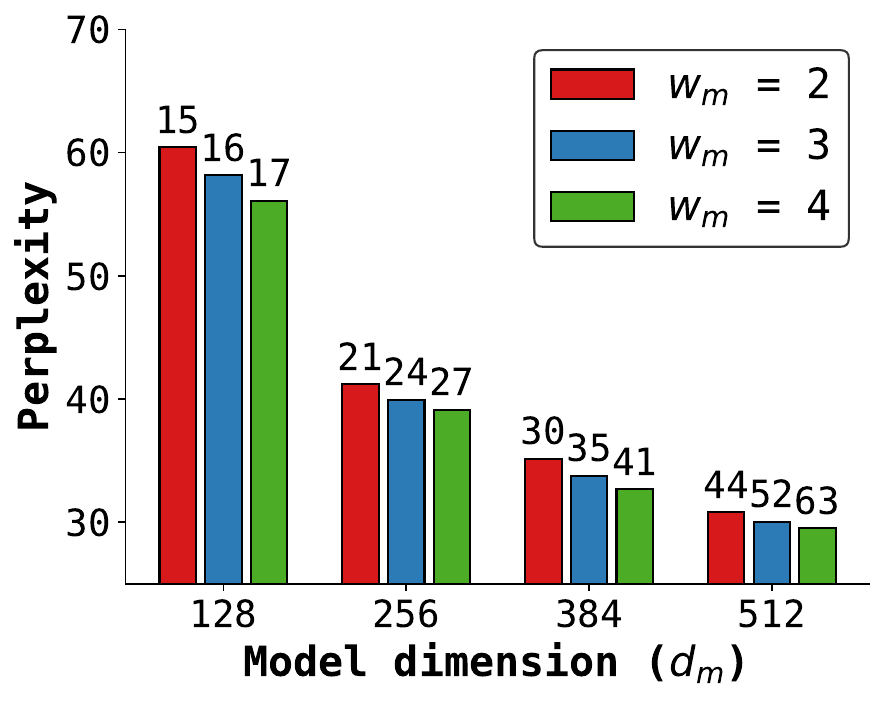}
            \caption{$N_{min}$=4, $N_{max}$=12}
            \label{fig:arch_4_12}
        \end{subfigure}
        \hfill
         \begin{subfigure}[b]{0.49\columnwidth}
            \centering
            \includegraphics[width=0.9\columnwidth]{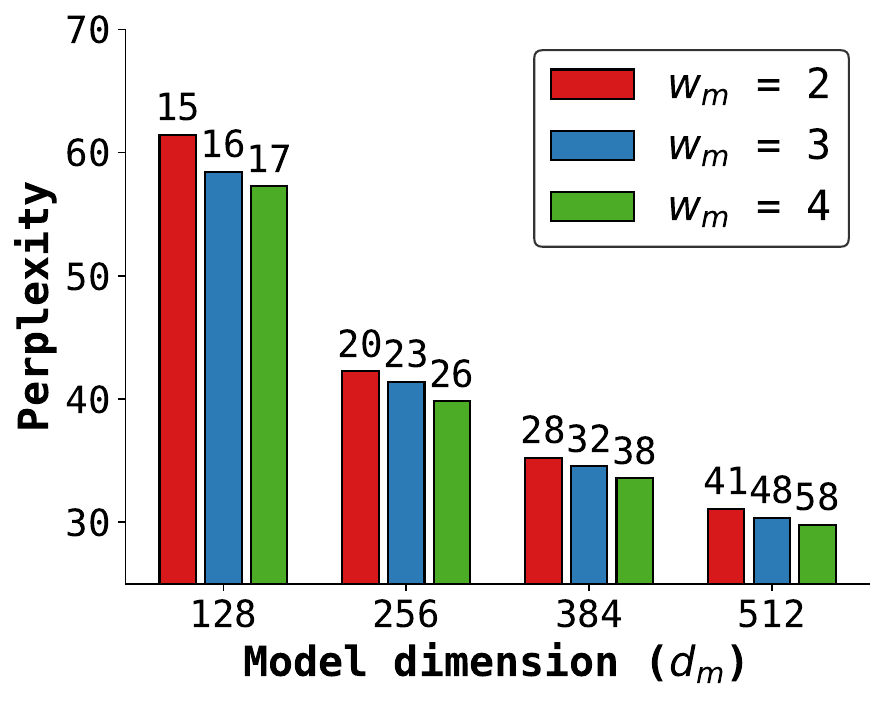}
            \caption{$N_{min}$=6, $N_{max}$=12}
            \label{fig:arch_6_12}
        \end{subfigure}
        \caption{\textbf{Scaling up \arch}. Scaling one configuration parameter (e.g., $d_m$) while keeping other configuration parameters constant (e.g., $N_{min}$, $N_{max}$, and $w_m$) consistently improves performance. The numbers on top of each bar represents network parameters (in million). Lower value of perplexity means better performance.}
        \label{fig:appendix_delight_scaling}
\end{figure}

Table \ref{tab:scaling_ablte} studies the impact of \arch~block parameters on the WikiText-103 dataset, namely (1) minimum number of GLTs $N_{min}$, (2) maximum number of GLTs $N_{max}$, (3) width multiplier $w_m$, and (4) model dimension $d_m$ (see Figure \ref{fig:redefine_transformer_sa}). Figure~\ref{fig:dextra_define}, Figure~\ref{fig:shuffling}, and Figure~\ref{fig:reduction_fac} shows the impact of the \dextra, feature shuffling, and the light-weight FFN. Table \ref{tab:dextra_position} shows the effect of position of \dextra~in the \arch~block while Figure \ref{fig:appendix_delight_scaling} shows the effect of scaling \arch~networks. We choose the WikiText-103 dataset for ablations because it has very large vocabulary compared to other datasets (267K vs. 30-40K), allowing us to test the ability under large vocabulary sizes. The performance is reported in terms of perplexity (lower is better) on the validation set. In our ablation studies, we used the same settings for training as in Section \ref{ssec:language_mdoeling} except that we train only for 50K iterations.

\noindent{\bf DeLighT block:} Overall, Table~\ref{tab:scaling_ablte} shows that scaling depth and width using \dextra~and block-wise scaling improves performance. We make following observations: 
\vspace{-1mm}
\begin{enumerate}[leftmargin=*, label=\alph*)]
\setlength\itemsep{0em}
\item Block-wise scaling (R4, R5) delivers better performance compared to uniform scaling (R1-R3). For instance, \arch~with $N_{min}=4$ and $N_{max}=8$ (R4) is $1.25\times$ shallower than \arch~with $N_{min}=8$ and $N_{max}=8$ (R2), but delivers better performance with a similar  number of parameters and operations. Scaling $w_m$ improves performance (R2 vs. R3), however, the improvement is significantly lower than for the model with block-wise scaling (R3 vs. R5). This suggests that non-uniform distribution of parameters across blocks allows the network to learn better representations.  

\item Different ratios between $N_{max}$ and $N_{min}$ yields different results. We observe significant performance improvements when the ratio is greater than or equal to two. For example, when we scale $\frac{N_{max}}{N_{min}}$ from 2 to 3 (R6 vs. R8), the perplexity improves by ${\sim}5$ points with only a moderate increase in network parameters. On the other hand, when the $\frac{N_{max}}{N_{min}}$ is close to 1 (R6 vs. R7), performance does not change appreciably. This is likely because the allocation of parameters across blocks is close to uniform (Eq. \ref{eq:mw}). This is consistent with our previous observation. 

\item Learning shallower and narrower representations near the input and deeper and wider representations near the output achieves better performance. For example, when we scaled $N_{max}$ from 8 to 12 for $N_{min}=4$ (R6, R8), \arch~delivered better performance with a similar number of parameters compared to a model with $N_{min}=6$ (R7, R9). This is likely because the ratio of $N_{max}$ and $N_{min}$ is higher when $N_{min}=4$, which helps allocate parameters per block more effectively. 
\item Deeper and wider representations near the input and shallower and narrower representations near the output hurts performance (R13 vs. R16). 
\item Scaling width using $w_m$ and $d_m$ improves performance (R10-R15), however, their impact is different. For example, when we scale $w_m$ and $d_m$ by two, the rate of increase in number of parameters and operations is more rapid with  $d_m$ compared to $w_m$. \arch's~ability to learn wider representations in different ways may be useful in selecting application specific models.
\end{enumerate}

\begin{table}[t!]
    \centering
        \resizebox{0.8\columnwidth}{!}{
            \begin{tabular}{l|cccc|cccc}
              \toprule[1.5pt]
              \rowcolor{white}
              \textbf{Row \#} & $N_{min}$  & $N_{max}$ & $w_m$ & $d_m$ & \textbf{Depth} & \textbf{Parameters}  & \textbf{MACs} & \textbf{Perplexity} \\
              \toprule[1.25pt]
              \multicolumn{9}{c}{\textbf{Uniform vs. block-wise scaling}} \\
              \midrule
                R1 & 4 & 4 & 2 & 256 & 43 & 14.1 M & 2.96 B  & 56.19 \\
                R2 & 8 & 8 & 2 & 256 & 115 & 16.6 M & 3.49 B & 48.58 \\
                R3 & 8 & 8 & 4 & 256 & 115 & 22.1 M & 4.64 B & 45.10 \\
                \cmidrule[0.25pt]{2-9}
                R4 & 4 & 8 & 2 & 256 & 92 & 16.7 M & 3.51 B & 46.30 \\
                R5 & \cellcolor{red!25} 4 & \cellcolor{red!25} 12 & \cellcolor{red!25} 2 & \cellcolor{red!25} 256 & 158 & 21.0 M & 4.41 B & 41.18 \\
              \toprule[1pt]
              \multicolumn{9}{c}{\textbf{Varying depth ($N_{min}$ and $N_{max}$ (Eq. \ref{eq:mw})}} \\
              \midrule
                R6 & 4 & 8 & 2 & 256 & 92 & 16.7 M & 3.51 B & 46.30 \\
                R7 & 6 & 8 & 2 & 256 & 102 & 16.5 M & 3.46 B & 46.68 \\
                R8 & \cellcolor{red!25} 4 & \cellcolor{red!25} 12 & \cellcolor{red!25} 2 & \cellcolor{red!25} 256 & 158 & 21.0 M & 4.41 B & 41.18 \\
                R9 & 6 & 12 & 2 & 256 & 172 & 20.0 M & 4.20 B & 42.26  \\
              \toprule[1pt]
              \multicolumn{9}{c}{\textbf{Varying \dextra's width $w_m$ (Eq. \ref{eq:mw})}} \\
              \midrule
                R10 & \cellcolor{red!25} 4 & \cellcolor{red!25} 12 & \cellcolor{red!25} 2 & \cellcolor{red!25} 256 & 158 & 21.0 M & 4.41 B & 41.18 \\
                R11 & 4 & 12 & 3 & 256 & 158 & 23.8 M & 4.99 B & 39.92  \\
                R12 & 4 & 12 & 4 & 256 & 158 & 27.1 M & 5.69 B & 39.10 \\
              \toprule[1pt]
              \multicolumn{9}{c}{\textbf{Varying model width $d_m$}} \\
              \midrule 
                R13 & \cellcolor{red!25} 4 & \cellcolor{red!25} 12 & \cellcolor{red!25} 2 & \cellcolor{red!25} 256 & 158 &  21.0 M & 4.41 B & 41.18 \\
                R14 & 4 & 12 & 2 & 384 & 158 & 29.9 M & 6.28 B & 35.14 \\
                R15 & 4 & 12 & 2 & 512 & 158 & 43.8 M & 9.20 B & 30.81 \\
              \toprule[1pt]
              \multicolumn{9}{c}{\textbf{Deeper and wider near the Input}} \\
              \midrule 
               R16 & 12 &  4 & 2 &  256 & 158 &  21.0 M & 4.41 B & 43.10 \\
            \bottomrule[1.5pt]
            \end{tabular}
        }
    \caption{\textbf{Ablations on different aspects of the \arch~block}, including uniform vs. block-wise scaling, depth scaling, and width scaling. Rows partially highlighted in color have the same configuration (repeated for illustrating results). Our experimental setup is similar to Section \ref{sec:results}, except that we train our models for 50K iterations. Multiplication and addition operations (MACs) are computed for 20 time steps. }
    \label{tab:scaling_ablte}
\end{table}

\vspace{1mm}
\noindent{\bf Impact of \dextra:} We replace \dextra~in the \arch~block (Figure \ref{fig:redefine_transformer_sa}) with (1) the DeFINE transformation and (2) a stack of linear layers. Figure \ref{fig:dextra_define} shows that \dextra~delivers similar performance with significantly fewer parameters compared to the DeFINE unit and linear layers. In these experiments, the settings are the same as R13-R15 (Table \ref{tab:scaling_ablte}), except, $N_{max}=8$, because models with a stack of linear layers learn too many parameters.

\vspace{1mm}
\noindent{\bf Feature shuffling:} Figure \ref{fig:shuffling} shows that feature shuffling improves the performance of \arch~by 1-2 perplexity points. Here, we use the same settings as in R13-R15 (Table \ref{tab:scaling_ablte}).

\vspace{1mm}
\noindent{\bf Light-weight FFN:} Figure \ref{fig:reduction_fac} shows the impact of varying the reduction factor $r$ in the light-weight FFN. We use the same settings as in R13 (Table \ref{tab:scaling_ablte}). We did not observe any significant drop in performance until $r=4$. Beyond $r=4$, we see a drop in performance (perplexity increases by ${\sim}2$ points). In such cases, the inner dimensions of the light-weight FFN are very small and hurt performance. Notably, the light-weight FFN with $r=2^2$ delivered the same performance as $r=2^{-2}$, but with $1.28\times$ fewer network parameters. At $r=2^{-2}$, the light-weight FFN is the same as the FFN in \cite{vaswani2017attention}. This suggests that the ability of \dextra~to learn representations in high-dimensional spaces efficiently allows us to reduce the computational burden on the FFN. 

We also tested removing the light-weight FFN and while it reduced parameters by $\sim$0.5-1 M, performance dropped by about 2-3 perplexity points across different parametric settings.

\begin{figure}[t!]
    \centering
    \begin{subfigure}[b]{0.5\columnwidth}
        \centering
        \resizebox{!}{120px}{
            \input{tikz/scaling_new.tikz}\scaling
        }
        \caption{}
    \end{subfigure}
    \hfill
    \begin{subfigure}[b]{0.48\columnwidth}
        \includegraphics[width=\columnwidth]{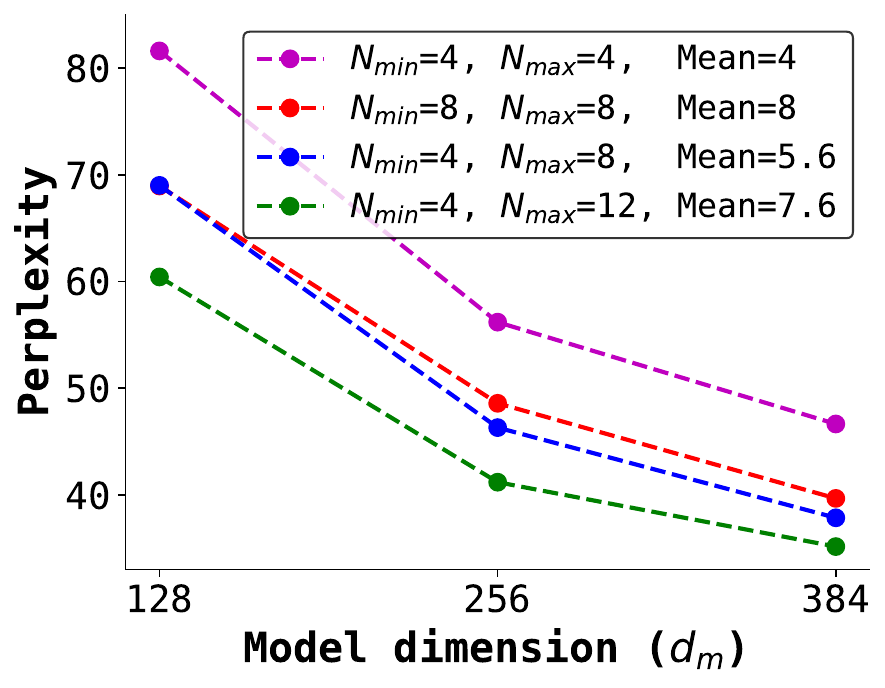}
        \caption{}
    \end{subfigure}
    \caption{\textbf{Uniform vs. block-wise scaling.} (a) contrasts the uniform and block-wise scaling methods. (b) compares the results of \arch~with uniform and block-wise scaling methods on the WikiText-103 dataset. \arch~networks with block-wise scaling delivers better performance across different settings. Lower perplexity value means better performance.}
    \label{fig:appendix_block_unifrom}
\end{figure}

\vspace{1mm}
\noindent{\bf Uniform vs. block-wise scaling:} Figure \ref{fig:appendix_block_unifrom} compares the performance of \arch~with uniform and block-wise scaling. For a given model dimension $d_m$, \arch~models with block-wise scaling delivers better performance.

\paragraph{Position of \dextra:}
%\label{ssec:ablate_position}
We studied three configurations for the \dextra~on the WikiText-103 validation set (Table \ref{tab:dextra_position}): (1) \dextra~followed by single-headed attention and light-weight FFN, (2) single-headed attention followed by \dextra, and (3) single-headed attention followed by \dextra~and light-weight FFN. For similar number of parameters, we found that (2) and (3) drops the performance of (1) significantly across different parametric settings. This suggests that deeper and wider representations helps learn better contextual representations; allowing us to replace multi-headed attention with single-headed attention.

\begin{table}[t!]
    \centering
    \resizebox{0.9\columnwidth}{!}{
    \begin{tabular}{lrr}
        \toprule[1.5pt]
        \textbf{Configuration}  & \textbf{Parameters} & \textbf{Perplexity} \\
        \midrule[1pt]
        \dextra + Single-head attention + Light-weight FFN & 31 M & \textbf{34.20} \\ 
        Single-head attention + \dextra & 30 M & 39.02 \\ 
        Single-head attention + \dextra + Light-weight FFN & 31 M & 39.43 \\ 
        \midrule
        \dextra + Single-head attention + Light-weight FFN & 99 M & \textbf{23.16} \\ 
        Single-head attention + \dextra & 96 M & 28.33 \\ 
        Single-head attention + \dextra + Light-weight FFN & 99 M & 27.94 \\
        \bottomrule[1.5pt]
    \end{tabular}
    }
    \caption{\textbf{Effect of the position of \dextra}. Lower value of perplexity means better performance.}
    \label{tab:dextra_position}
\end{table}

\vspace{1mm}
\noindent{\bf Scaling up \arch:} Figure \ref{fig:appendix_delight_scaling} shows the results of \arch~models obtained after varying configuration parameters of \arch~transformations ($N_{min}$=\{4, 6\}, $N_{max}$=\{8, 12\}, $w_m$=\{2, 3, 4\}, and $d_m$=\{256, 384, 512\}). We can see that scaling one configuration parameter (e.g., $d_m$) while keeping other configuration parameters constant (e.g., $N_{min}$, $N_{max}$, and $w_m$) consistently improves performance.

This work investigates relationships between $N_{min}$, $N_{max}$, $w_m$, and $d_m$, manually. We believe that a more principled approach, such as compound scaling of \citet{tan2019efficientnet}, that establishes relationships between these parameters would produce more efficient and accurate models. %We will explore such methods in the future.

\section{Source code for group linear transformation}
\label{sec:append_source}
The source code for implementing group linear transformation (GLT) in PyTorch is shown in Listing \ref{lst:navie_glt}. The source code for efficiently implementing the grouping function in GLT is shown in Listing \ref{lst:group_kernel}. Since the ungrouping kernel is similar to grouping kernel, we have not shown it here.

The reshape and transpose operations in naive PyTorch implementation for grouping and ungrouping are replaced with a dedicated CUDA kernels, resulting in reduced memory footprint and faster training.

\definecolor{dkgreen}{rgb}{0,0.6,0}
\definecolor{gray}{rgb}{0.5,0.5,0.5}
\definecolor{mauve}{rgb}{0.58,0,0.82}

\lstset{frame=H,
  language=Python,
  aboveskip=3mm,
  belowskip=3mm,
  showstringspaces=false,
  columns=flexible,
  basicstyle={\small\ttfamily},
  numbers=none,
  numberstyle=\tiny\color{gray},
  keywordstyle=\color{blue},
  commentstyle=\color{dkgreen},
  stringstyle=\color{mauve},
  breaklines=true,
  breakatwhitespace=true,
  tabsize=3
}

\begin{minipage}{\columnwidth}
\begin{lstlisting}[caption="Naive implementation of GLT in Pytorch", label={lst:navie_glt}]
import torch
def glt_function(x, n_groups, weights, bias=None):
    '''
    :param x: Input tensor of size [B x N], where B is batch size and N is input dimension
    :param n_groups: number of groups in GLT
    :param weights: glt weights [g x N/g x M/g]
    :param bias: GLT bias (optional) of size [g x 1 x M/g]
    :return: output tensor of size [B x M]
    '''
    bsz = x.size(0)
    
    ## GROUPING FUNCTION: Converts [B x N] tensor to [g x B  x N/g] ##
    # [B x N] --> [B x g  x N/g]
    x = x.contiguous().view(bsz, n_groups, -1)
    # [B x g x N/g] --> [g x B  x N/g]
    x = x.transpose(0, 1)  # transpose so that group is first

    ## TRANSFORMATION FUNCTION: Transforms from N/g-dimensional space to M/g-dimensional space ##
    # [g x B  x N/g] x [g x N/g x M/g] --> [g x B x M/g]
    x = torch.bmm(x, weights)  # multiply with Weights
    # add bias
    if bias is not None:
        x = torch.add(x, bias)
        
    ## REGROUPING FUNCTION: Converts [g x B x M/g] tensor to [B x M] ##
    # [g x B x M/g] --> [B x g x M/g]
    x = x.transpose(0, 1)  # transpose so that batch is first
    # [B x g x M/g] --> [B x M]
    x = x.contiguous().view(bsz, -1)
    return x
\end{lstlisting}
\end{minipage}

\lstset{language=C++}
\begin{minipage}{\columnwidth}
\begin{lstlisting}[caption="Grouping kernel in CUDA", label={lst:group_kernel}]

/* Grouping Kernel: Transforms input from [B x N] to [g x B x N/g] */
template<typename scalar_t>
__global__ void grouping_kernel_forward(const scalar_t* input,
                            const int groups, const int total_elements,
                            const int input_features, const int group_features,
                            const int batch_size, scalar_t* output){
    const int index = IMUL(blockIdx.x, blockDim.x) + threadIdx.x;
    if (index >= total_elements){
        return;
    }
    const int b_idx = index / group_features;
    const int g_f_idx = (index % group_features);
    int in_offset, out_offset;
    #pragma unroll
    for(int g=0; g <  groups; g++){
        in_offset = (b_idx * input_features) + (g * group_features) + g_f_idx;
        out_offset = ((g * batch_size + b_idx) * group_features) + g_f_idx;
        output[out_offset] = input[in_offset];
    }
}
\end{lstlisting}
\end{minipage}

\end{document}

%% file: tikz/self_attention_units.tikz
\newcommand\lw{1mm}
\tikzset{>=triangle 45}

\newcommand{\drawTrapezoidFFN}[1]{
        \path[draw=black, fill=cyan!60, line width=\lw] (#1) -- ($(#1) + (6, 0)$) -- ($(#1) + (8, -2)$) -- ($(#1) + (-2, -2)$) -- (#1);
        
        \path[draw=black, fill=cyan!60, line width=\lw] ($(#1) + (-2, -4)$) -- ($(#1) + (8, -4)$)-- ($(#1) + (6, -6)$) -- ($(#1) + (0, -6)$) -- ($(#1) + (-2, -4)$);
        
        \draw[thick, ->, line width=\lw] ($(#1) + (3, -2)$) -- ($(#1) + (3, -4)$);
        \node [] at ($(#1) + (4.5, 1)$) {\scalebox{4}{$d_m$}};
        \node [] at ($(#1) + (5.75, -3)$) {\scalebox{4}{$d_f$=$4d_m$}};
        \node [] at ($(#1) + (4.5, -7)$) {\scalebox{4}{ $d_m$}};
}

\newcommand{\drawTrapezoidReduce}[1]{
        \path[draw=black, fill=cyan!60, line width=\lw] (#1) -- ($(#1) + (6, 0)$) -- ($(#1) + (4, -2)$) -- ($(#1) + (2, -2)$) -- (#1);
}

\newcommand{\drawTrapezoidFlat}[1]{
        \path[draw=black, fill=cyan!60, line width=\lw] (#1) -- ($(#1) + (6, 0)$) -- ($(#1) + (6, -2)$) -- ($(#1) + (0, -2)$) -- (#1);
}

\definecolor{mygreen}{RGB}{51,160,44}

\newcommand{\transformer}{
    \begin{tikzpicture}[block/.style={
inner sep=8,outer sep=0,
align=center,rounded corners, line width=\lw,
font=\fontsize{35}{60}\selectfont}]

    \node (x) at (0, 1) {};
    
    %Keys
    \drawTrapezoidReduce{$(x) + (0.6, -1.6)$};
    \drawTrapezoidReduce{$(x) + (0.3, -1.8)$};
    \drawTrapezoidReduce{$(x) + (0, -2)$};
    \node [] at ($(x) + (3, -3)$) {\fontsize{35}{20}\selectfont Key};
    \node [] at ($(x) + (4.5, -0.75)$) {\scalebox{4}{$d_m$}};
    \node [] at ($(x) + (4.5, -5)$) {\scalebox{4}{$d_h$}};
    
    %Query
    \drawTrapezoidReduce{$(x) + (-6.4, -1.6)$};
    \drawTrapezoidReduce{$(x) + (-6.7, -1.8)$};
    \drawTrapezoidReduce{$(x) + (-7, -2)$};
    \node [] at ($(x) + (-4, -3)$) {\fontsize{35}{20}\selectfont Query};
    \node [] at ($(x) + (-5.5, -0.75)$) {\scalebox{4}{$d_m$}};
    \node [] at ($(x) + (-5.5, -5)$) {\scalebox{4}{$d_h$}};
    
     %Value
    \drawTrapezoidReduce{$(x) + (7.6, -1.6)$};
    \drawTrapezoidReduce{$(x) + (7.3, -1.8)$};
    \drawTrapezoidReduce{$(x) + (7, -2)$};
    \node [] at ($(x) + (10, -3)$) {\fontsize{35}{20}\selectfont Value};
    \node [] at ($(x) + (11.5, -0.75)$) {\scalebox{4}{$d_m$}};
    \node [] at ($(x) + (11.5, -5)$) {\scalebox{4}{$d_h$}};

    % draw attention blocks
    \node [block, draw, fill=white, minimum height=2cm, minimum width=6cm] at ($(x) + (3.6, -7.6)$) {};
    \node [block, draw, fill=white, minimum height=2cm, minimum width=6cm] at ($(x) + (3.3, -7.8)$) {};
    \node [block, draw, fill=white, minimum height=2cm, minimum width=6cm]  (attn1) at ($(x) + (3, -8)$) {\fontsize{35}{20}\selectfont Attention};
    
    % draw connections to K, Q, V
    \draw[thick, ->, line width=\lw] ($(x) + (3, 1)$) -- ($(x) + (3, -2)$);
    \draw[thick, ->, line width=\lw] ($(x) + (3, 0.25)$) -| ($(x) + (10, -2)$);
    \draw[thick, ->, line width=\lw] ($(x) + (3, 0.25)$) -| ($(x) + (-4, -2)$);
    
    % draw connections to Attention
    \draw[thick, ->, line width=\lw] ($(x) + (3, -4)$) -- (attn1);
    \draw[thick, ->, line width=\lw] ($(x) + (10, -4)$) |- (attn1);
    \draw[thick, ->, line width=\lw] ($(x) + (-4, -4)$) |- (attn1);
    
    % draw linear layer
    \node [block, draw, minimum height=2cm, minimum width=4cm] (concat) at ($(attn1) + (0, -3.5)$) {Concat};
    
    \drawTrapezoidFlat{$(concat) + (-3, -3)$};
    \node [] at ($(concat) + (1.5, -2)$) {\scalebox{4}{$d_m$}};
    \node [] at ($(concat) + (1.5, -6)$) {\scalebox{4}{$d_m$}};
    
    % draw add
    \node [block, draw, minimum height=2cm, minimum width=4cm] (add) at ($(concat) + (0, -8)$) {Add};
    
    % draw connections
    \draw[thick, ->, line width=\lw] (attn1) -- (concat);
    \draw[thick, ->, line width=\lw] (concat) -- ($(concat) + (0, -3)$);
    \draw[thick, ->, line width=\lw] ($(concat) + (0, -5)$) -- (add);
    \draw[thick, ->, line width=\lw] ($(x) + (3, 0.25)$) -- ($(x) + (14, 0.25)$) |- (add);

    % draw FFN
    \drawTrapezoidFFN{$(add) + (-3, -3)$};
    \node [block, draw, minimum height=2cm, minimum width=3cm] (add1) at ($(add) + (0, -12)$) {Add};
    %\node [block, draw, minimum height=2cm, minimum width=4cm, right=2cm of add1, fill=cyan] (add2) {};
    %\node [block, minimum height=2cm, minimum width=4cm, right=0.5cm of add2] (add3) {Linear};
    %\drawTrapezoidExpand{$(add) + (-3, -8)$};
    
    % draw connections
    \draw[thick, ->, line width=\lw] (add) -- ($(add) + (-6.5, 0)$) |- (add1);
    \draw[thick, ->, line width=\lw] (add) -- ($(add) + (0, -3)$);
    \draw[thick, ->, line width=\lw] ($(add) + (0, -9)$) -- ($(add) + (0, -11)$);
    \draw[thick, ->, line width=\lw] ($(add1) + (0, -1)$) -- ($(add1) + (0, -2.5)$);
    
    % draw box around 
    
    \draw [decorate,decoration={calligraphic brace,amplitude=25pt,raise=8pt, mirror},yshift=0pt, line width=2*\lw]
($(x)+(-8, 0)$) -- ($(x)+(-8, -19.4)$) node [black,midway,xshift=-2cm, rotate=90] {\scalebox{3.5}{Multi-head Attention}};

    \draw [decorate,decoration={calligraphic brace,amplitude=25pt,raise=8pt, mirror},yshift=0pt, line width=2*\lw]
($(x)+(-8, -19.7)$) -- ($(x)+(-8, -33)$) node [black,midway,xshift=-2cm, rotate=90] {\scalebox{3.5}{Feed Forward Network (FFN)}};

    \node [] at ($(x)+(-4, -12)$) {\fontsize{40}{20}\selectfont \textcolor{red}{Attention ops:}};
    
    \node [] at ($(x)+(-4, -14)$) {{\scalebox{5}{\textcolor{red}{$\mathcal{O}(d_m n^2)$}}}};
    
    \node [] at ($(x)+(11, -23)$) {\fontsize{40}{20}\selectfont \textcolor{red}{FFN params:}};
    
    \node [] at ($(x)+(12, -25)$) {{\scalebox{5}{\textcolor{red}{$8d_m^2$}}}};
    
    \node [text width=8cm, fill=yellow!40, minimum height=3.5cm, align=center, right=1.75cm of add1, rounded corners=5mm, draw=black, line width=\lw] {\fontsize{40}{20}\selectfont \text{Depth = 4}};
    
    \end{tikzpicture}
}

\newcommand{\drawDextra}[1]{
        \path[draw=black, fill=green!30, line width=\lw] (#1) -- ($(#1) + (6, 0)$) -- ($(#1) + (8, -2)$) -- ($(#1) + (5, -4)$) -- ($(#1) + (1, -4)$) -- ($(#1) + (-2, -2)$) -- (#1);
        %\node [] (add1) at ($(#1) + (3, -3)$) {\fontsize{35}{20}\selectfont \dextra};
        \node [] at ($(#1) + (1.5, 1.5)$) {\scalebox{4}{$d_m$}};
        %\node [] () at () {};
        \draw [decorate,decoration={brace,amplitude=15pt,mirror,raise=8pt},yshift=0pt, line width=1.5*\lw]($(#1) + (-2.25, 0)$) -- ($(#1) + (-2.25, -4)$) node [black,midway,xshift=-2.5cm, yshift=0.75cm]{{\scalebox{5}{$N^b$}}};
         \node [] at ($(#1) + (-5.5, -3.5)$) {\scalebox{4}{(Eq. \ref{eq:mw})}};
        
        \draw [<->, line width=1.5\lw] ($(#1) + (-2, -2)$) -- ($(#1) + (8, -2)$);
        %\node [] (add1) at ($(#1) + (3, -1)$) {\scalebox{3.5}{Width}};
        \node [] (add1) at ($(#1) + (3, -1)$) {\scalebox{4}{$w_m^b d_m$}};
}

\newcommand{\drawTrapezoidFlatA}[1]{
        \path[draw=black, fill=cyan!60, line width=\lw] (#1) -- ($(#1) + (4, 0)$) -- ($(#1) + (4, -2)$) -- ($(#1) + (0, -2)$) -- (#1);
}

\newcommand{\drawProj}[1]{
        \path[draw=black, fill=cyan!60, line width=\lw] (#1) -- ($(#1) + (4, 0)$) -- ($(#1) + (6, -2)$) -- ($(#1) + (-2, -2)$) -- (#1);
}

\newcommand{\drawLightFFN}[1]{
        \path[draw=black, fill=cyan!60, line width=\lw] (#1) -- ($(#1) + (6, 0)$) -- ($(#1) + (4, -2)$) -- ($(#1) + (2, -2)$) -- (#1);
        
        \path[draw=black, fill=cyan!60, line width=\lw] ($(#1) + (2, -4)$) -- ($(#1) + (4, -4)$)-- ($(#1) + (6, -6)$) -- ($(#1) + (0, -6)$) -- ($(#1) + (2, -4)$);
        
        \draw[thick, ->, line width=\lw] ($(#1) + (3, -2)$) -- ($(#1) + (3, -4)$);
        \node [] at ($(#1) + (5, 1)$) {\scalebox{4}{ $d_m$}};
        \node [] at ($(#1) + (5.5, -3)$) {\scalebox{4}{$d_m/4$}};
        \node [] at ($(#1) + (5, -7)$) {\scalebox{4}{ $d_m$}};
}

\newcommand{\redefine}{
    \begin{tikzpicture}[block/.style={
inner sep=8,outer sep=0,
align=center,rounded corners, line width=\lw,
font=\fontsize{35}{60}\selectfont}]

    \node (x) at (0, 1) {};
    
    \drawDextra{$(x) + (0, -2)$};
    
    %Keys
    \drawTrapezoidFlatA{$(x) + (1, -9)$};
    \node [] at ($(x) + (3, -10)$) {\fontsize{35}{20}\selectfont Key};
    \node [] at ($(x) + (4, -8)$) {\scalebox{4}{$d_o$}};
    \node [] at ($(x) + (5.5, -12)$) {\scalebox{4}{$d_o$=$\frac{d_m}{2}$}};
    
    %Query
    \drawTrapezoidFlatA{$(x) + (-5.6, -9)$};
    \node [] at ($(x) + (-3.5, -10)$) {\fontsize{35}{20}\selectfont Query};
    \node [] at ($(x) + (-2.35, -8)$) {\scalebox{4}{$d_o$}};
    \node [] at ($(x) + (-2.35, -12)$) {\scalebox{4}{$d_o$}};
    
     %Value
    \drawTrapezoidFlatA{$(x) + (7.6, -9)$};
    \node [] at ($(x) + (9.5, -10)$) {\fontsize{35}{20}\selectfont Value};
    \node [] at ($(x) + (11, -8)$) {\scalebox{4}{$d_o$}};
    \node [] at ($(x) + (11, -12)$) {\scalebox{4}{$d_o$}};
    
    % draw attention
    \node [block, draw, fill=white, minimum height=2cm, minimum width=6cm]  (attn1) at ($(x) + (3, -15)$) {\fontsize{35}{20}\selectfont Attention};
    
    \drawProj{$(attn1) + (-2, -3)$};
    \node [] at ($(attn1) + (1.5, -2)$) {\scalebox{4}{$d_o$}};
    \node [] at ($(attn1) + (1.5, -6)$) {\scalebox{4}{$d_m$}};
    
    % draw add
    \node [block, draw, minimum height=2cm, minimum width=4cm] (add) at ($(attn1) + (0, -8)$) {Add};
    
    %draw light FFN
    \drawLightFFN{$(attn1) + (-3, -11)$};
    
    \node [block, draw, minimum height=2cm, minimum width=4cm] (add1) at ($(attn1) + (0, -20)$) {Add};

    % draw connections
    \draw[thick, ->, line width=\lw] ($(x) + (3, 0)$) -- ($(x) + (3, -2)$);
    
    \draw[thick, ->, line width=\lw] ($(x) + (3, -6)$) -- ($(x) + (3, -9)$);
    \draw[thick, ->, line width=\lw] ($(x) + (3, -7)$) -| ($(x) + (9.6, -9)$);
    \draw[thick, ->, line width=\lw] ($(x) + (3, -7)$) -| ($(x) + (-3.6, -9)$);
    
    \draw[thick, ->, line width=\lw] ($(x) + (3, -11)$) -- (attn1);
    \draw[thick, ->, line width=\lw] ($(x) + (9.6, -11)$) |- (attn1);
    \draw[thick, ->, line width=\lw] ($(x) + (-3.6, -11)$) |- (attn1);
    
    \draw[thick, ->, line width=\lw] (attn1) -- ($(attn1) + (0, -3)$);
    
    \draw[thick, ->, line width=\lw] ($(attn1) + (0, -5)$) -- (add);
    \draw[thick, ->, line width=\lw] ($(x) + (3, -1)$) --  ($(x) + (13, -1)$) |- (add);
    
    % draw connections
    \draw[thick, ->, line width=\lw] (add) -- ($(add) + (-5, 0)$) |- (add1);
    \draw[thick, ->, line width=\lw] (add) -- ($(add) + (0, -3)$);
    \draw[thick, ->, line width=\lw] ($(add) + (0, -9)$) -- ($(add) + (0, -11)$);
    \draw[thick, ->, line width=\lw] ($(add1) + (0, -1)$) -- ($(add1) + (0, -2.5)$);
    
    % decoration
    \draw [decorate,decoration={calligraphic brace,amplitude=25pt,raise=8pt, mirror},yshift=0pt, line width=2*\lw]
($(x)+(-8, 0)$) -- ($(x)+(-8, -23)$) node [black,midway,xshift=-2.25cm, rotate=90] {\scalebox{3.5}{\dextra~with Single-head Attention}};

    \draw [decorate,decoration={calligraphic brace,amplitude=25pt,raise=8pt, mirror},yshift=0pt, line width=2*\lw]
($(x)+(-8, -23.7)$) -- ($(x)+(-8, -37)$) node [black,midway,xshift=-2.25cm, rotate=90] {\scalebox{3.5}{Light-weight FFN}};

    \node [] at ($(x)+(-4, -17)$) {\fontsize{40}{20}\selectfont \textcolor{red}{Attention ops:}};
    
    \node [] at ($(x)+(-4.25, -19)$) {{\scalebox{5}{\textcolor{red}{$\mathcal{O}(d_o n^2)$}}}};
    
    \node [] at ($(x)+(11, -27)$) {\fontsize{40}{20}\selectfont \textcolor{red}{FFN params:}};
    
    \node [] at ($(x)+(11, -29.25)$) {{\scalebox{5.5}{\textcolor{red}{$\frac{d_m^2}{2}$}}}};

    \node [text width=9.25cm, fill=yellow!40, minimum height=3.5cm, align=center, right=1.75cm of add1, rounded corners=5mm, draw=black, line width=\lw] {\fontsize{40}{20}\selectfont \text{Depth = 4 + }\scalebox{1.25}{$N^b$}};

    \end{tikzpicture}
}

%% file: tikz/dextra_new.tex
\newcommand\lw{1mm}
\newcommand\rishift{5}
\tikzset{>=triangle 45}

\tikzstyle{node_box} = [rectangle,  fill=white, font=\fontsize{25}{25}\selectfont, rounded corners=0.5mm, line width=\lw, align=center]

\tikzstyle{enode} = [rectangle, draw, line width=\lw, minimum height=1.5cm, text width=10cm, align=center, rounded corners=0.75mm, font=\fontsize{25}{25}\selectfont,]

\definecolor{mygreen}{RGB}{51,160,44}

\newcommand{\define}{
    \begin{tikzpicture}[block/.style={
inner sep=8,outer sep=0,
align=center,rounded corners, line width=\lw}]

    \node (x) at (0, 0) {};
    
    % 8 GROUPs
    \path[draw=black, fill=red!20, line width=\lw] (x.north) -- ($(x.north) + (1, 0)$) -- ($(x.north) + (0.5, -2)$) -- ($(x.north) + (-2, -2)$) -- (x.north);
    \path[draw=black, fill=black!20, line width=\lw] ($(x.north) + (1, 0)$) -- ($(x.north) + (2, 0)$) -- ($(x.north) + (1, -2)$) -- ($(x.north) + (-0.5, -2)$) -- ($(x.north) + (1, 0)$);
    \path[draw=black, fill=blue!40, line width=\lw] ($(x.north) + (2, 0)$) -- ($(x.north) + (3, 0)$) -- ($(x.north) + (2.5, -2)$) -- ($(x.north) + (1, -2)$) -- ($(x.north) + (2, 0)$);
    \path[draw=black, fill=orange!40, line width=\lw] ($(x.north) + (3, 0)$) -- ($(x.north) + (4, 0)$) -- ($(x.north) + (4, -2)$) -- ($(x.north) + (2.5, -2)$) -- ($(x.north) + (3, 0)$);
    \path[draw=black, fill=red!80, line width=\lw] ($(x.north) + (4, 0)$) -- ($(x.north) + (5, 0)$) -- ($(x.north) + (5.5, -2)$) -- ($(x.north) + (4, -2)$) -- ($(x.north) + (4, 0)$);
    \path[draw=black, fill=black!60, line width=\lw] ($(x.north) + (5, 0)$) -- ($(x.north) + (6, 0)$) -- ($(x.north) + (7, -2)$) -- ($(x.north) + (5.5, -2)$) -- ($(x.north) + (5, 0)$);
    \path[draw=black, fill=blue!80, line width=\lw] ($(x.north) + (6, 0)$) -- ($(x.north) + (7, 0)$) -- ($(x.north) + (8.5, -2)$) -- ($(x.north) + (7, -2)$) -- ($(x.north) + (6, 0)$);
    \path[draw=black, fill=orange!60, line width=\lw] ($(x.north) + (7, 0)$) -- ($(x.north) + (8, 0)$) -- ($(x.north) + (10, -2)$) -- ($(x.north) + (8.5, -2)$) -- ($(x.north) + (7, 0)$);

    % 4 GROUPS
    \path[draw=black, fill=red!20, line width=\lw] ($(x.north) + (-2, -4)$) -- ($(x.north) + (1, -4)$) -- ($(x.north) + (0, -6)$) -- ($(x.north) + (-4, -6)$) -- ($(x.north) + (-2, -4)$);
    \path[draw=black, fill=black!20, line width=\lw] ($(x.north) + (1, -4)$) -- ($(x.north) + (4, -4)$) -- ($(x.north) + (4, -6)$) -- ($(x.north) + (0, -6)$) -- ($(x.north) + (1, -4)$);
    \path[draw=black, fill=blue!40, line width=\lw] ($(x.north) + (4, -4)$) -- ($(x.north) + (7, -4)$) -- ($(x.north) + (8, -6)$) -- ($(x.north) + (4, -6)$) -- ($(x.north) + (4, -4)$);
    \path[draw=black, fill=orange!40, line width=\lw] ($(x.north) + (7, -4)$) -- ($(x.north) + (10, -4)$) -- ($(x.north) + (12, -6)$) -- ($(x.north) + (8, -6)$) -- ($(x.north) + (7, -4)$);
    
    % 2 GROUPS
    \path[draw=black, fill=red!20, line width=\lw] ($(x.north) + (-4, -8)$) -- ($(x.north) + (4, -8)$) -- ($(x.north) + (4, -10)$) -- ($(x.north) + (-6, -10)$) -- ($(x.north) + (-4, -8)$);
    \path[draw=black, fill=black!20, line width=\lw] ($(x.north) + (4, -8)$) -- ($(x.north) + (12, -8)$) -- ($(x.north) + (14, -10)$) -- ($(x.north) + (4, -10)$) -- ($(x.north) + (4, -8)$);
    
    % 1 group
    \path[draw=black, fill=red!20, line width=\lw] ($(x.north) + (-6, -12)$) -- ($(x.north) + (14, -12)$) -- ($(x.north) + (16, -14)$) -- ($(x.north) + (-8, -14)$) -- ($(x.north) + (-6, -12)$);
    
    % reduce
    \path[draw=black, fill=red!20, line width=\lw] ($(x.north) + (-8, -16)$) -- ($(x.north) + (16, -16)$) -- ($(x.north) + (6, -18)$) -- ($(x.north) + (2, -18)$) -- ($(x.north) + (-8, -16)$);

    % draw connections
    \node (l1) at ($(x)+(4, 3)$) {\scalebox{6}{Input ($d_m$-dimensional)}};
    \draw[thick, ->, line width=2*\lw] ($(x) + (4, 2)$) -- ($(x) + (4, 0)$);
    \draw[thick, ->, line width=2*\lw] ($(x) + (4, -2)$) -- ($(x) + (4, -4)$);
    \draw[thick, ->, line width=2*\lw] ($(x) + (4, -6)$) -- ($(x) + (4, -8)$);
    \draw[thick, ->, line width=2*\lw] ($(x) + (4, -10)$) -- ($(x) + (4, -12)$);
    \draw[thick, ->, line width=2*\lw] ($(x) + (4, -14)$) -- ($(x) + (4, -16)$);
    \draw[thick, ->, line width=2*\lw] ($(x) + (4, -18)$) -- ($(x) + (4, -20)$);
    \node (l1) at ($(x)+(4, -21)$) {\scalebox{6}{Output ($d_o$-dimensional)}};
    
    \draw [decorate,decoration={calligraphic brace,amplitude=20pt,raise=6pt, mirror},yshift=0pt, line width=2*\lw]
(-9, 3) -- (-9, -14) node [black,midway,xshift=-2.5cm, rotate=90] {\scalebox{6}{Expansion}};

    \draw [decorate,decoration={calligraphic brace,amplitude=20pt,raise=6pt, mirror},yshift=0pt, line width=2*\lw]
(-9, -14.5) -- (-9, -21) node [black,midway,xshift=-2.5cm, rotate=90] {\scalebox{6}{ Reduction}};

    \draw [decorate,decoration={calligraphic brace,amplitude=20pt,raise=6pt},yshift=0pt, line width=2*\lw]
(17, 3) -- (17, -21) node [black,midway,xshift=2.5cm, rotate=270] {\scalebox{6}{ No. of layers (depth) = $N$}};

    \end{tikzpicture}
}

\newcommand{\invHGTShuffle}{
    \begin{tikzpicture}[block/.style={
inner sep=8,outer sep=0,
align=center,rounded corners, line width=\lw}]

    \node (x) at (0, 0) {};
    
    % 1 GROUPs
    \path[draw=black, fill=red!20, line width=\lw] (x.north) -- ($(x.north) + (8, 0)$) -- ($(x.north) + (10.67, -2)$) -- ($(x.north) + (-2.67, -2)$) -- (x.north);
    
    % 2 GROUPs
    \path[draw=black, fill=red!20, line width=\lw] ($(x.north) + (-2.67, -4)$) -- ($(x.north) + (4, -4)$) -- ($(x.north) + (4, -6)$) -- ($(x.north) + (-5.34, -6)$) -- ($(x.north) + (-2.67, -4)$) ;
    \path[draw=black, fill=black!20, line width=\lw] ($(x.north) + (4, -4)$) -- ($(x.north) + (10.67, -4)$) -- ($(x.north) + (13.34, -6)$) -- ($(x.north) + (4, -6)$) -- ($(x.north) + (4, -4)$) ;
    
    % 4 GROUPS
    \path[draw=black, fill=red!20, line width=\lw] ($(x.north) + (-5.34, -8)$) -- ($(x.north) + (-0.64, -8)$) -- ($(x.north) + (-2, -10)$) -- ($(x.north) + (-8, -10)$) -- ($(x.north) + (-5.34, -8)$) ;
    \path[draw=black, fill=black!20, line width=\lw] ($(x.north) + (-0.64, -8)$) -- ($(x.north) + (4, -8)$) -- ($(x.north) + (4, -10)$) -- ($(x.north) + (-2, -10)$) -- ($(x.north) + (-0.64, -8)$) ;
    \path[draw=black, fill=blue!40, line width=\lw] ($(x.north) + (4, -8)$) -- ($(x.north) + (8.7, -8)$) -- ($(x.north) + (10, -10)$) -- ($(x.north) + (4, -10)$) -- ($(x.north) + (4, -8)$) ;
    \path[draw=black, fill=orange!40, line width=\lw] ($(x.north) + (8.7, -8)$) -- ($(x.north) + (13.34, -8)$) -- ($(x.north) + (16, -10)$) -- ($(x.north) + (10, -10)$) -- ($(x.north) + (8.7, -8)$) ;
    
    % 2 GROUPS
    \path[draw=black, fill=red!20, line width=\lw] ($(x.north) + (-8, -12)$) -- ($(x.north) + (4, -12)$) -- ($(x.north) + (4, -14)$) -- ($(x.north) + (-5.34, -14)$) -- ($(x.north) + (-8, -12)$) ;
    \path[draw=black, fill=black!20, line width=\lw] ($(x.north) + (4, -12)$) -- ($(x.north) + (16, -12)$) -- ($(x.north) + (13.34, -14)$) -- ($(x.north) + (4, -14)$) -- ($(x.north) + (4, -12)$) ;
    
    %1 GROUP
    \path[draw=black, fill=red!20, line width=\lw] ($(x.north) + (-5.34, -16)$) -- ($(x.north) + (13.34, -16)$) -- ($(x.north) + (6, -18)$) -- ($(x.north) + (2, -18)$) -- ($(x.north) + (-5.34, -16)$);
    
    % draw connections
    \node (l1) at ($(x)+(4, 3)$) {\scalebox{6}{Input ($d_m$-dimensional)}};
    \draw[thick, ->, line width=2*\lw] ($(x) + (4, 2)$) -- ($(x) + (4, 0)$);
    \draw[thick, ->, line width=2*\lw] ($(x) + (4, -2)$) -- ($(x) + (4, -4)$);
    \draw[thick, ->, line width=2*\lw] ($(x) + (4, -6)$) -- ($(x) + (4, -8)$);
    \draw[thick, ->, line width=2*\lw] ($(x) + (4, -10)$) -- ($(x) + (4, -12)$);
    \draw[thick, ->, line width=2*\lw] ($(x) + (4, -14)$) -- ($(x) + (4, -16)$);
    \draw[thick, ->, line width=2*\lw] ($(x) + (4, -18)$) -- ($(x) + (4, -20)$);
    \node (l1) at ($(x)+(4, -21)$) {\scalebox{6}{Output ($d_o$-dimensional)}};
    
    \draw [decorate,decoration={calligraphic brace,amplitude=20pt,raise=6pt, mirror},yshift=0pt, line width=2*\lw]
(-9, 3) -- (-9, -10) node [black,midway,xshift=-2.5cm, rotate=90] {\scalebox{6}{ Expansion}};

    \draw [decorate,decoration={calligraphic brace,amplitude=20pt,raise=6pt, mirror},yshift=0pt, line width=2*\lw]
(-9, -10.5) -- (-9, -21) node [black,midway,xshift=-2.5cm, rotate=90] {\scalebox{6}{ Reduction}};

    \draw [decorate,decoration={calligraphic brace,amplitude=20pt,raise=.5pt},yshift=0pt, line width=2*\lw]
(17, 3) -- (17, -21) node [black,midway,xshift=2.5cm, rotate=270] {\scalebox{6}{ No. of layers (depth) = $N$}};

    \end{tikzpicture}
}

%% file: tikz/scaling_new.tikz
\newcommand\lw{0.5mm}
\tikzset{>=triangle 45}

\definecolor{cyanmy}{RGB}{0, 255, 255}
\definecolor{redmat}{RGB}{229, 0, 0}

\newcommand{\drawTrap}[3]{
        \path[draw=black, fill=cyanmy, line width=\lw] (#1) -- ($(#1)+(1, 0)$) -- ($(#1)+(1+#2, -#3)$) -- ($(#1)+(1, -2*#3)$) -- ($(#1)+(0, -2 * #3)$) -- ($(#1)+(-#2, -#3)$) -- (#1);
}

\newcommand{\drawRect}[1]{
        \path[draw=black, fill=redmat, line width=\lw] (#1) -- ($(#1) + (2, 0)$) -- ($(#1) + (2, -2)$) -- ($(#1) + (0, -2)$) -- (#1);
}

\newcommand{\scaling}{
    \begin{tikzpicture}
    
    \node (a) at (2, 3) {\scalebox{4}{\bf Block-wise}};
    \node (b) at (-8.25, 3) {\scalebox{4}{\bf Uniform}};
    \node (inp) at (0.5, 1.5) {\scalebox{3}{Input}};
    \drawTrap{$(0, -0.75)$}{0.25}{0.25};    
    \drawTrap{$(0, -3.6)$}{0.5}{0.5};    
    \drawTrap{$(0, -7)$}{1}{1};
    \node (out) at (0.5, -10.5) {\scalebox{3}{Output}};
    
	\draw[thick, ->, line width=\lw] (inp) -- (0.5, -0.75);
	\draw[thick, ->, line width=\lw] (0.5, -1.25) -- (0.5, -3.6);
	\draw[thick, ->, line width=\lw, dashed] (0.5, -4.6) -- (0.5, -7);
	\draw[thick, ->, line width=\lw, dashed] (0.5, -9) -- (out);
	
	\node (inp) at (-9, 1.5) {\scalebox{3}{Input}};
	\drawRect{$(-10, 0)$};
    \drawRect{$(-10, -3.1)$};
    \drawRect{$(-10, -7)$};
    \node (out) at (-9, -10.5) {\scalebox{3}{Output}};
    
    \draw[thick, ->, line width=\lw] (inp) -- (-9, 0);
    \draw[thick, ->, line width=\lw] (-9, -2) -- (-9, -3.1);
	\draw[thick, ->, line width=\lw, dashed] (-9, -5.1) -- (-9, -7);
	\draw[thick, ->, line width=\lw, dashed] (-9, -9) -- (out);
	
	 \draw [decorate,decoration={calligraphic brace,amplitude=14pt,raise=4pt, mirror},yshift=0pt, line width=1.5*\lw]
(-11.5, 0) -- (-11.5, -9) node [black,midway,xshift=-1.5cm, rotate=90] {\scalebox{4}{$\mathcal{B}$\ blocks}};

    \node [black] at (5.5, -1) {\scalebox{3.75}{$N^0$ = $N_{min}$}};
    \node [black] at (6.5, -8) {\scalebox{3.75}{$N^{\mathcal{B}-1}$ = $N_{max}$}};

    \node [fill=white] at (5.75, -5) {\scalebox{4}{\textcolor{blue}{(see Eq. \ref{eq:mw})}}};

    \node [black] at (-5, -8) {\scalebox{3.75}{$N^{\mathcal{B}-1}$=$N$}};
    \node [black] at (-5.5, -1) {\scalebox{3.75}{$N^0$=$N$}};

	% draw box
     
     \path[draw=black, line width=1.5*\lw] (-13.75, 4) -- (10.5, 4) -- (10.5, -11.25) -- (-13.75, -11.25) -- (-13.75, 4);
    
   	\end{tikzpicture}
}  

%% file: tikz/enc_dec.tikz
\newcommand\lw{0.45mm}
\tikzset{>=triangle 45}

%'Transformer-base': {215, 25, 28},
%        'ReDeFINE': {44, 123, 182}

\definecolor{tam}{RGB}{215, 25, 28} %{251,128,114}
\definecolor{ram}{RGB}{44, 123, 182}%227,26,28}

\tikzstyle{enode} = [rectangle, draw, line width=\lw, minimum height=0.8cm, text width=2cm, align=center, rounded corners=0.5mm]

\tikzset{%
  do path picture/.style={%
    path picture={%
      \pgfpointdiff{\pgfpointanchor{path picture bounding box}{south west}}%
        {\pgfpointanchor{path picture bounding box}{north east}}%
      \pgfgetlastxy\x\y%
      \tikzset{x=\x/2,y=\y/2}%
      #1
    }
  },
  sin wave/.style={do path picture={    
    \draw [line cap=round] (-3/4,0)
      sin (-3/8,1/2) cos (0,0) sin (3/8,-1/2) cos (3/4,0);
  }},
  cross/.style={do path picture={    
    \draw [line cap=round] (-1,-1) -- (1,1) (-1,1) -- (1,-1);
  }},
  plus/.style={do path picture={    
    \draw [line cap=round] (-3/4,0) -- (3/4,0) (0,-3/4) -- (0,3/4);
  }}
}

\newcommand{\drawDextra}[1]{
        \path[draw=black, fill=green!40, line width=\lw] (#1) -- ($(#1) + (1, 0)$) -- ($(#1) + (1.5, -0.5)$) -- ($(#1) + (1, -1)$) -- ($(#1) + (0, -1)$) -- ($(#1) + (-0.5, -0.5)$) -- (#1);
        \node [text width=1cm, align=center] (add1) at ($(#1) + (0.5, -0.5)$) { };
}

\newcommand{\encdecattn}{
    \begin{tikzpicture}
    
    \node (inp) at (0, -1) {Inputs};
    \node[enode, fill=red!20] (emb) at (0, 0.5) {Look-up Table};
	\node[minimum height=1cm] (emb-r) at (0, 2) {};  
	\drawDextra{$(emb-r) + (-0.5, 0.5)$};
	
	\node [circle, draw, sin wave, line width=\lw, minimum size=1cm] (emb-p) at (-1.75, 3.75) {}; % posistional embedding
	\node [circle, draw, plus, line width=\lw, minimum size=0.75cm] (emb-add) at (0, 3.75) {};
	\node [left=0.01cm of emb-p, text width=1.5cm] {Positional Encoding};
	
	% draw connections
	\draw [->, thick, line width=\lw] (inp) -- (emb);
	\draw [->, thick, line width=\lw] (emb) -- (emb-r);
	\draw [->, thick, line width=\lw] (emb-r) -- (emb-add);
    \draw [->, thick, line width=\lw] (emb-p) -- (emb-add);
    
    % box around first two layers
    \draw [-, dashed, line width=2*\lw, rounded corners, color=blue] (-1.5, -0.5) -- (-1.5, 2.75) -- (1.5, 2.75) -- (1.5, -0.5) -- (-1.5, -0.5);
    \node [] (n) at (-2.75, 1.5) {\fontsize{14}{20}\selectfont \textcolor{blue}{Embedding}};
    \node [below=0.001cm of n] {\fontsize{14}{20}\selectfont \textcolor{blue}{Layer}};

    % draw the attention unit
     \node[minimum height=1cm] (attn-red) at (0, 6) {};  
	 \drawDextra{$(attn-red) + (-0.5, 0.5)$};
	  %\node[enode, fill=green!20] (attn-red) at (0, 6) {\dextra}; 
	  \node[left=0.5cm of attn-red] (temp) {};
	  %\node[above=0.001cm of temp] {\fontsize{15}{50}\selectfont $N^{L_i}$};
	  %\node[below=0.001cm of temp] {\fontsize{15}{50}\selectfont $m_w^{L_i}$};
	  
      \node[enode, fill=orange!40] (attn-attn) at (0, 8) {Single-head Attention};
      \node[enode, fill=yellow!40] (attn-norm) at (0, 9.5) {Add \& Norm};
      \node[enode, fill=cyan!40] (attn-ffn) at (0, 11.5) {Light-weight FFN};
      \node[enode, fill=yellow!40] (attn-norm1) at (0, 13.25) {Add \& Norm};

     % draw connections
     \draw [->, thick, line width=\lw] (emb-add) -- (attn-red);
     \draw [->, thick, line width=\lw] (attn-red) -- (attn-attn);
     \draw [->, thick, line width=\lw] (attn-attn) -- (attn-norm);
     \draw [->, thick, line width=\lw] (attn-norm) -- (attn-ffn);
     \draw [->, thick, line width=\lw] (attn-ffn) -- (attn-norm1);
     
     \draw [rounded corners, ->, thick, line width=\lw] (0, 6.85) -- (1, 6.85) -- (1, 7.6);
     \draw [rounded corners, ->, thick, line width=\lw] (0, 6.85) -- (-1, 6.85) -- (-1, 7.6);
     \draw [rounded corners, ->, thick, line width=\lw] (0, 5) -- (-2.5, 5) -- (-2.5, 9.5) -- (attn-norm);
     \draw [rounded corners, ->, thick, line width=\lw] (0, 10.25) -- (-2.5, 10.25) -- (-2.5, 13.25) -- (attn-norm1);
     
     % draw box around ATTN
     \draw [-, dashed, line width=2*\lw, rounded corners, color=blue] (-3, 4.75) -- (-3, 14) -- (1.5, 14) -- (1.5, 4.75) -- (-3, 4.75);
     \node [rotate=90, color=blue] (ne) at (-3.5, 9.25) {\fontsize{20}{50}\selectfont $\mathcal{B} \times$};

     % draw the decoder
    \node (inpd) at (5, -1) {Outputs (shifted right)};
    \node[enode, fill=red!20] (embd) at (5, 0.5) {Look-up Table};
	%\node[enode, fill=green!20] (embd-r) at (5, 2) {\dextra};    
	\node[minimum height=1cm] (embd-r) at (5, 2) {};  
	 \drawDextra{$(embd-r) + (-0.5, 0.5)$};
	
	\node [circle, draw, sin wave, line width=\lw, minimum size=1cm] (embd-p) at (6.75, 3.75) {}; % posistional embedding
	\node [circle, draw, plus, line width=\lw, minimum size=0.75cm] (embd-add) at (5, 3.75) {};
	\node [right=0.01cm of embd-p, text width=1.5cm] {Positional Encoding};
	
	% draw connections
	\draw [->, thick, line width=\lw] (inpd) -- (embd);
	\draw [->, thick, line width=\lw] (embd) -- (embd-r);
	\draw [->, thick, line width=\lw] (embd-r) -- (embd-add);
    \draw [->, thick, line width=\lw] (embd-p) -- (embd-add);
    
    % box around first two layers
    \draw [-, dashed, line width=2*\lw, rounded corners, color=blue] (-1.5+5, -0.5) -- (-1.5+5, 2.75) -- (1.5+5, 2.75) -- (1.5+5, -0.5) -- (-1.5+5, -0.5);
    %\node [rotate=90, color=red] (nd) at (-2+5, 1.0) {Output Embedding};
    \node [] (nd) at (-2.5+5+5.25, 1.75) {\fontsize{14}{20}\selectfont \textcolor{blue}{Embedding}};
    \node [below=0.01cm of nd] {\fontsize{14}{20}\selectfont \textcolor{blue}{Layer}};
    
    %\node[enode, fill=green!20] (attnd-red) at (0+5, 6) {\dextra}; 
    \node[minimum height=1cm] (attnd-red) at (0+5, 6) {};  
	 \drawDextra{$(attnd-red) + (-0.5, 0.5)$};
    \node[right=0.5cm of attnd-red] (temp) {};
	%\node[above=0.001cm of temp] {\fontsize{15}{50}\selectfont $N^{L_i}$};
	%\node[below=0.001cm of temp] {\fontsize{15}{50}\selectfont $m_w^{L_i}$};
    \node[enode, fill=orange!40, text width=2.5cm] (attnd-attn) at (0+5, 8) {Masked Single-head Attention};
      \node[enode, fill=yellow!40] (attnd-norm) at (0+5, 9.5) {Add \& Norm};
      
    %\node[enode, fill=black!20] (attnd-lin) at (-1.1+5, 11.5) {Linear};
    %\node[enode, fill=green!20] (attnd-red2) at (5, 11.5) {\dextra};
    
    %\node[minimum height=1cm] (attnd-red2) at (5, 11.5) {};  
	% \drawDextra{$(attnd-red2) + (-0.5, 0.5)$};
    
    %\node[right=0.5cm of attnd-red2] (temp) {};
	%\node[above=0.001cm of temp] {\fontsize{15}{50}\selectfont $N^{L_i}$};
	%\node[below=0.001cm of temp] {\fontsize{15}{50}\selectfont $m_w^{L_i}$};
	
      \node[enode, fill=orange!40] (attnd-attn1) at (0+5, 11.5) {Single-head Attention};
      \node[enode, fill=yellow!40] (attnd-norm1) at (0+5, 13.25) {Add \& Norm};
      \node[enode, fill=cyan!40] (attnd-ffn) at (0+5, 15) {Light-weight FFN};
      \node[enode, fill=yellow!40] (attnd-norm2) at (0+5, 16.75) {Add \& Norm};
      
      \node[enode, fill=black!20] (cls) at (0+5, 18.25) {Linear};
      \node[enode] (cls1) at (0+5, 20) {Softmax};
      \node[] (cls2) at (0+5, 21.25) {Logits};
      
     % draw connections
	    \draw [rounded corners, ->, thick, line width=\lw] (0+5, 6.85) -- (1+5, 6.85) -- (1+5, 7.6);
     \draw [rounded corners, ->, thick, line width=\lw] (0+5, 6.85) -- (-1+5, 6.85) -- (-1+5, 7.6);     
     
     \draw [rounded corners,->, thick, line width=\lw] (embd-add) -- (attnd-red);
     \draw [rounded corners,->, thick, line width=\lw] (attnd-red) -- (attnd-attn);
     \draw [rounded corners,->, thick, line width=\lw] (attnd-attn) -- (attnd-norm);
     \draw [rounded corners,->, thick, line width=\lw] (attnd-norm) -- (attnd-attn1);
     \draw [rounded corners,->, thick, line width=\lw] (attn-norm1) -- (0, 14.5) -- (2.15, 14.5) -- (2.15, 10.25) -- (4.1, 10.25) -- (4.1, 11);
     \draw [rounded corners,->, thick, line width=\lw] (attn-norm1) -- (0, 14.5) -- (2.15, 14.5) -- (2.15, 10.25) -- (4.5, 10.25) -- (4.5, 11);
     
     %\draw [rounded corners,->, thick, line width=\lw] (attnd-lin) -- (5-1.1, 12.25) -- (3.25, 12.25) -- (3.25, 12.9);
     %\draw [rounded corners,->, thick, line width=\lw] (attnd-lin) -- (5-1.1, 12.25) -- (4.5, 12.25) -- (4.5, 12.9);
     
     %\draw [rounded corners,->, thick, line width=\lw] (attnd-red2) -- (attnd-attn1);
     \draw [rounded corners,->, thick, line width=\lw] (attnd-attn1) -- (attnd-norm1);
     \draw [rounded corners,->, thick, line width=\lw] (attnd-norm1) -- (attnd-ffn);
     \draw [rounded corners,->, thick, line width=\lw] (attnd-ffn) -- (attnd-norm2);
     
     % draw residual connections
     \draw [rounded corners, ->, thick, line width=\lw] (0+5, 5) -- (7.75, 5) -- (7.75, 9.5) -- (attnd-norm);
     \draw [rounded corners,->, thick, line width=\lw] (attnd-norm) -- (5, 10.4) -- (7.75, 10.4) -- (7.75, 13.25) -- (attnd-norm1);
     
     \draw [rounded corners,->, thick, line width=\lw] (attnd-norm1) -- (0+5, 14) --(7.75, 14) -- (7.75, 16.75) -- (attnd-norm2);
     
     % draw box around ATTN
     \draw [-, dashed, line width=2*\lw,rounded corners, color=blue] (-3+6, 4.75) -- (-3+6, 19) -- (1.5+5.5+1+0.5, 19) -- (1.5+5.5+1+0.5, 4.75) -- (-3+6, 4.75);
     \node [rotate=90, color=blue](ne) at (-3.5+12+0.5, 12) {\fontsize{20}{50}\selectfont $\mathcal{B} \times$};
     
     \draw [rounded corners,->, thick, line width=\lw] (attnd-norm2) -- (cls);
     \draw [rounded corners,->, thick, line width=\lw] (cls) -- (cls1);
     \draw [rounded corners,->, thick, line width=\lw] (cls1) -- (cls2);

     % draw weight tying
     \draw [rounded corners,<->, dashed, thick, line width=2*\lw, color=red] (embd) -- (emb);
     \draw [rounded corners,<->, dashed, thick, line width=2*\lw, color=red] (embd) --(9.75, 0.5) -- (9.75, 18.25) -- (cls);
     \node [rotate=90, color=red] (nd) at (10.25, 10) {\fontsize{14}{20}\selectfont Input and output weights are tied};

    \end{tikzpicture}
}

\newcommand{\decattn}{
    \begin{tikzpicture}
    
     % draw the decoder
    \node (inpd) at (5, -1) {Inputs (shifted right)};
    \node[enode, fill=red!20] (embd) at (5, 0.5) {Adaptive Inputs};
	%\node[enode, fill=green!20] (embd-r) at (5, 2) {\dextra};   
	\node[minimum height=1cm] (embd-r) at (5, 2) {};  
	 \drawDextra{$(embd-r) + (-0.5, 0.5)$};
	
	\node [circle, draw, sin wave, line width=\lw, minimum size=1cm] (embd-p) at (6.75, 3.75) {}; % posistional embedding
	\node [circle, draw, plus, line width=\lw, minimum size=0.75cm] (embd-add) at (5, 3.75) {};
	\node [right=0.01cm of embd-p, text width=1.5cm] {Positional Encoding};
	
	% draw connections
	\draw [->, thick, line width=\lw] (inpd) -- (embd);
	\draw [->, thick, line width=\lw] (embd) -- (embd-r);
	\draw [->, thick, line width=\lw] (embd-r) -- (embd-add);
    \draw [->, thick, line width=\lw] (embd-p) -- (embd-add);
    
    % box around first two layers
    \draw [-, dashed, line width=2*\lw, rounded corners, color=blue] (-1.5+5, -0.5) -- (-1.5+5, 2.75) -- (1.5+5, 2.75) -- (1.5+5, -0.5) -- (-1.5+5, -0.5);
    %\node [rotate=90, color=red] (nd) at (-2+5, 1.0) {Output Embedding};
    \node [] (nd) at (-2.5+5+5.25, 1.75) {\fontsize{14}{20}\selectfont \textcolor{blue}{Embedding}};
    \node [below=0.01cm of nd] {\fontsize{14}{20}\selectfont \textcolor{blue}{Layer}};
    
    %\node[enode, fill=green!20] (attnd-red) at (0+5, 6) {\dextra}; 
    \node[minimum height=1cm] (attnd-red) at (0+5, 6) {};  
	 \drawDextra{$(attnd-red) + (-0.5, 0.5)$};
    
    \node[right=0.5cm of attnd-red] (temp) {};
	%\node[above=0.001cm of temp] {\fontsize{15}{50}\selectfont $N^{L_i}$};
	%\node[below=0.001cm of temp] {\fontsize{15}{50}\selectfont $m_w^{L_i}$};
    \node[enode, fill=orange!40, text width=2.5cm] (attnd-attn) at (0+5, 8) {Masked Single-head Attention};
      \node[enode, fill=yellow!40] (attnd-norm) at (0+5, 9.5) {Add \& Norm};
      
      \node[enode, fill=cyan!40] (attnd-ffn) at (0+5, 11.5) {Light-weight FFN};
      \node[enode, fill=yellow!40] (attnd-norm2) at (0+5, 13.25) {Add \& Norm};
      
      \node[enode, fill=black!20] (cls) at (0+5, 15) {Adaptive Softmax};
      \node[] (cls2) at (0+5, 16.25) {Logits};
      
     % draw connections
	    \draw [rounded corners, ->, thick, line width=\lw] (0+5, 6.85) -- (1+5, 6.85) -- (1+5, 7.6);
     \draw [rounded corners, ->, thick, line width=\lw] (0+5, 6.85) -- (-1+5, 6.85) -- (-1+5, 7.6);     
     
     \draw [rounded corners,->, thick, line width=\lw] (embd-add) -- (attnd-red);
     \draw [rounded corners,->, thick, line width=\lw] (attnd-red) -- (attnd-attn);
     \draw [rounded corners,->, thick, line width=\lw] (attnd-attn) -- (attnd-norm);
     \draw [rounded corners,->, thick, line width=\lw] (attnd-norm) -- (attnd-ffn);
     \draw [rounded corners,->, thick, line width=\lw] (attnd-ffn) -- (attnd-norm2);
     
     % draw residual connections
     \draw [rounded corners, ->, thick, line width=\lw] (0+5, 5) -- (7.75, 5) -- (7.75, 9.5) -- (attnd-norm);
     \draw [rounded corners,->, thick, line width=\lw] (attnd-norm) -- (5, 10.4) -- (7.75, 10.4) -- (7.75, 13.25) -- (attnd-norm2);
     
     %\draw [rounded corners,->, thick, line width=\lw] (attnd-norm1) -- (0+5, 15.5) --(7.75, 15.5) -- (7.75, 18.25) -- (attnd-norm2);
     
     % draw box around ATTN
     \draw [-, dashed, line width=2*\lw,rounded corners, color=blue] (-3+6, 4.75) -- (-3+6, 14) -- (1.5+5.5+1+0.5, 14) -- (1.5+5.5+1+0.5, 4.75) -- (-3+6, 4.75);
     \node [rotate=90, color=blue](ne) at (-3.5+12+0.5, 9) {\fontsize{20}{50}\selectfont $\mathcal{B} \times$};
     
     \draw [rounded corners,->, thick, line width=\lw] (attnd-norm2) -- (cls);
     \draw [rounded corners,->, thick, line width=\lw] (cls) -- (cls2);

     % draw weight tying
     \draw [rounded corners,<->, dashed, thick, line width=2*\lw, color=red] (embd) --(9.75, 0.5) -- (9.75, 15) -- (cls);
     \node [rotate=90, color=red] (nd) at (10.25, 9) {\fontsize{14}{20}\selectfont Input and output weights are tied};

    \end{tikzpicture}
}

%% file: main_camera_ready.bbl
\begin{thebibliography}{50}
\providecommand{\natexlab}[1]{#1}
\providecommand{\url}[1]{\texttt{#1}}
\expandafter\ifx\csname urlstyle\endcsname\relax
  \providecommand{\doi}[1]{doi: #1}\else
  \providecommand{\doi}{doi: \begingroup \urlstyle{rm}\Url}\fi

\bibitem[Vaswani et~al.(2017)Vaswani, Shazeer, Parmar, Uszkoreit, Jones, Gomez,
  Kaiser, and Polosukhin]{vaswani2017attention}
Ashish Vaswani, Noam Shazeer, Niki Parmar, Jakob Uszkoreit, Llion Jones,
  Aidan~N Gomez, {\L}ukasz Kaiser, and Illia Polosukhin.
\newblock Attention is all you need.
\newblock In \emph{Advances in neural information processing systems}, pages
  5998--6008, 2017.

\bibitem[Raffel et~al.(2019)Raffel, Shazeer, Roberts, Lee, Narang, Matena,
  Zhou, Li, and Liu]{raffel2019exploring}
Colin Raffel, Noam Shazeer, Adam Roberts, Katherine Lee, Sharan Narang, Michael
  Matena, Yanqi Zhou, Wei Li, and Peter~J Liu.
\newblock Exploring the limits of transfer learning with a unified text-to-text
  transformer.
\newblock \emph{arXiv preprint arXiv:1910.10683}, 2019.

\bibitem[Brown et~al.(2020)Brown, Mann, Ryder, Subbiah, Kaplan, Dhariwal,
  Neelakantan, Shyam, Sastry, Askell, Agarwal, Herbert-Voss, Krueger, Henighan,
  Child, Ramesh, Ziegler, Wu, Winter, Hesse, Chen, Sigler, Litwin, Gray, Chess,
  Clark, Berner, McCandlish, Radford, Sutskever, and Amodei]{brown2020language}
Tom~B. Brown, Benjamin Mann, Nick Ryder, Melanie Subbiah, Jared Kaplan,
  Prafulla Dhariwal, Arvind Neelakantan, Pranav Shyam, Girish Sastry, Amanda
  Askell, Sandhini Agarwal, Ariel Herbert-Voss, Gretchen Krueger, Tom Henighan,
  Rewon Child, Aditya Ramesh, Daniel~M. Ziegler, Jeffrey Wu, Clemens Winter,
  Christopher Hesse, Mark Chen, Eric Sigler, Mateusz Litwin, Scott Gray,
  Benjamin Chess, Jack Clark, Christopher Berner, Sam McCandlish, Alec Radford,
  Ilya Sutskever, and Dario Amodei.
\newblock Language models are few-shot learners.
\newblock \emph{arXiv preprint arXiv:2005.14165}, 2020.

\bibitem[Devlin et~al.(2019)Devlin, Chang, Lee, and Toutanova]{devlin2018bert}
Jacob Devlin, Ming-Wei Chang, Kenton Lee, and Kristina Toutanova.
\newblock {BERT}: Pre-training of deep bidirectional transformers for language
  understanding.
\newblock In \emph{Proceedings of the 2019 Conference of the North {A}merican
  Chapter of the Association for Computational Linguistics: Human Language
  Technologies, Volume 1 (Long and Short Papers)}, 2019.

\bibitem[Hinton et~al.(2012)Hinton, Srivastava, Krizhevsky, Sutskever, and
  Salakhutdinov]{hinton2012improving}
Geoffrey~E Hinton, Nitish Srivastava, Alex Krizhevsky, Ilya Sutskever, and
  Ruslan~R Salakhutdinov.
\newblock Improving neural networks by preventing co-adaptation of feature
  detectors.
\newblock \emph{arXiv preprint arXiv:1207.0580}, 2012.

\bibitem[Wan et~al.(2013)Wan, Zeiler, Zhang, Le~Cun, and
  Fergus]{wan2013regularization}
Li~Wan, Matthew Zeiler, Sixin Zhang, Yann Le~Cun, and Rob Fergus.
\newblock Regularization of neural networks using dropconnect.
\newblock In \emph{International conference on machine learning}, pages
  1058--1066, 2013.

\bibitem[Merity et~al.(2018{\natexlab{a}})Merity, Keskar, and
  Socher]{merity2018regularizing}
Stephen Merity, Nitish~Shirish Keskar, and Richard Socher.
\newblock Regularizing and optimizing {LSTM} language models.
\newblock In \emph{International Conference on Learning Representations},
  2018{\natexlab{a}}.
\newblock URL \url{https://openreview.net/forum?id=SyyGPP0TZ}.

\bibitem[Mehta et~al.(2018)Mehta, Koncel-Kedziorski, Rastegari, and
  Hajishirzi]{mehta2018pyramidal}
Sachin Mehta, Rik Koncel-Kedziorski, Mohammad Rastegari, and Hannaneh
  Hajishirzi.
\newblock Pyramidal recurrent unit for language modeling.
\newblock In \emph{Proceedings of the 2018 Conference on Empirical Methods in
  Natural Language Processing}, 2018.

\bibitem[Zhang et~al.(2018)Zhang, Zhou, Lin, and Sun]{zhang2018shufflenet}
Xiangyu Zhang, Xinyu Zhou, Mengxiao Lin, and Jian Sun.
\newblock Shufflenet: An extremely efficient convolutional neural network for
  mobile devices.
\newblock In \emph{Proceedings of the IEEE conference on computer vision and
  pattern recognition}, pages 6848--6856, 2018.

\bibitem[Dai et~al.(2019)Dai, Yang, Yang, Carbonell, Le, and
  Salakhutdinov]{dai2019transformer}
Zihang Dai, Zhilin Yang, Yiming Yang, Jaime Carbonell, Quoc~V Le, and Ruslan
  Salakhutdinov.
\newblock Transformer-xl: Attentive language models beyond a fixed-length
  context.
\newblock In \emph{Association for Computational Linguistics}, 2019.

\bibitem[Child et~al.(2019)Child, Gray, Radford, and
  Sutskever]{child2019generating}
Rewon Child, Scott Gray, Alec Radford, and Ilya Sutskever.
\newblock Generating long sequences with sparse transformers.
\newblock \emph{arXiv preprint arXiv:1904.10509}, 2019.

\bibitem[Kitaev et~al.(2020)Kitaev, Kaiser, and Levskaya]{Kitaev2020Reformer}
Nikita Kitaev, Lukasz Kaiser, and Anselm Levskaya.
\newblock Reformer: The efficient transformer.
\newblock In \emph{International Conference on Learning Representations}, 2020.

\bibitem[Beltagy et~al.(2020)Beltagy, Peters, and Cohan]{Beltagy2020Longformer}
Iz~Beltagy, Matthew~E. Peters, and Arman Cohan.
\newblock Longformer: The long-document transformer.
\newblock \emph{arXiv:2004.05150}, 2020.

\bibitem[Raganato and Tiedemann(2018)]{raganato2018analysis}
Alessandro Raganato and J{\"o}rg Tiedemann.
\newblock An analysis of encoder representations in transformer-based machine
  translation.
\newblock In \emph{Proceedings of the 2018 {EMNLP} Workshop {B}lackbox{NLP}:
  Analyzing and Interpreting Neural Networks for {NLP}}, November 2018.

\bibitem[Brunner et~al.(2020)Brunner, Liu, Pascual, Richter, Ciaramita, and
  Wattenhofer]{Brunner2020On}
Gino Brunner, Yang Liu, Damian Pascual, Oliver Richter, Massimiliano Ciaramita,
  and Roger Wattenhofer.
\newblock On identifiability in transformers.
\newblock In \emph{International Conference on Learning Representations}, 2020.
\newblock URL \url{https://openreview.net/forum?id=BJg1f6EFDB}.

\bibitem[Voita et~al.(2019{\natexlab{a}})Voita, Sennrich, and
  Titov]{voita2019bottom}
Elena Voita, Rico Sennrich, and Ivan Titov.
\newblock The bottom-up evolution of representations in the transformer: A
  study with machine translation and language modeling objectives.
\newblock In \emph{Proceedings of the 2019 Conference on Empirical Methods in
  Natural Language Processing and the 9th International Joint Conference on
  Natural Language Processing (EMNLP-IJCNLP)}, 2019{\natexlab{a}}.

\bibitem[Michel et~al.(2019)Michel, Levy, and Neubig]{michel2019sixteen}
Paul Michel, Omer Levy, and Graham Neubig.
\newblock Are sixteen heads really better than one?
\newblock In \emph{Advances in Neural Information Processing Systems}, pages
  14014--14024, 2019.

\bibitem[Raganato et~al.(2020)Raganato, Scherrer, and
  Tiedemann]{raganato2020fixed}
Alessandro Raganato, Yves Scherrer, and J{\"o}rg Tiedemann.
\newblock Fixed encoder self-attention patterns in transformer-based machine
  translation.
\newblock \emph{arXiv preprint arXiv:2002.10260}, 2020.

\bibitem[Tay et~al.(2020)Tay, Bahri, Metzler, Juan, Zhao, and
  Zheng]{tay2020synthesizer}
Yi~Tay, Dara Bahri, Donald Metzler, Da-Cheng Juan, Zhe Zhao, and Che Zheng.
\newblock Synthesizer: Rethinking self-attention in transformer models.
\newblock \emph{arXiv preprint arXiv:2005.00743}, 2020.

\bibitem[Wu et~al.(2019)Wu, Fan, Baevski, Dauphin, and Auli]{wu2018pay}
Felix Wu, Angela Fan, Alexei Baevski, Yann Dauphin, and Michael Auli.
\newblock Pay less attention with lightweight and dynamic convolutions.
\newblock In \emph{International Conference on Learning Representations}, 2019.

\bibitem[Wu et~al.(2020)Wu, Liu, Lin, Lin, and Han]{Wu2020Lite}
Zhanghao Wu, Zhijian Liu, Ji~Lin, Yujun Lin, and Song Han.
\newblock Lite transformer with long-short range attention.
\newblock In \emph{International Conference on Learning Representations}, 2020.

\bibitem[So et~al.(2019)So, Le, and Liang]{so2019evolved}
David So, Quoc Le, and Chen Liang.
\newblock The evolved transformer.
\newblock In \emph{Proceedings of the 36th International Conference on Machine
  Learning}, pages 5877--5886, 2019.

\bibitem[Gehring et~al.(2017)Gehring, Auli, Grangier, Yarats, and
  Dauphin]{gehring2017convolutional}
Jonas Gehring, Michael Auli, David Grangier, Denis Yarats, and Yann~N Dauphin.
\newblock Convolutional sequence to sequence learning.
\newblock In \emph{Proceedings of the 34th International Conference on Machine
  Learning-Volume 70}, pages 1243--1252. JMLR. org, 2017.

\bibitem[Dauphin et~al.(2017)Dauphin, Fan, Auli, and
  Grangier]{dauphin2017language}
Yann~N Dauphin, Angela Fan, Michael Auli, and David Grangier.
\newblock Language modeling with gated convolutional networks.
\newblock In \emph{Proceedings of the 34th International Conference on Machine
  Learning-Volume 70}, pages 933--941. JMLR. org, 2017.

\bibitem[Lan et~al.(2020)Lan, Chen, Goodman, Gimpel, Sharma, and
  Soricut]{lan2020ALBERT}
Zhenzhong Lan, Mingda Chen, Sebastian Goodman, Kevin Gimpel, Piyush Sharma, and
  Radu Soricut.
\newblock Albert: A lite bert for self-supervised learning of language
  representations.
\newblock In \emph{International Conference on Learning Representations}, 2020.

\bibitem[Shoeybi et~al.(2019)Shoeybi, Patwary, Puri, LeGresley, Casper, and
  Catanzaro]{shoeybi2019megatron}
Mohammad Shoeybi, Mostofa Patwary, Raul Puri, Patrick LeGresley, Jared Casper,
  and Bryan Catanzaro.
\newblock Megatron-lm: Training multi-billion parameter language models using
  gpu model parallelism.
\newblock \emph{arXiv preprint arXiv:1909.08053}, 2019.

\bibitem[Tan and Le(2019)]{tan2019efficientnet}
Mingxing Tan and Quoc~V. Le.
\newblock Efficientnet: Rethinking model scaling for convolutional neural
  networks.
\newblock In Kamalika Chaudhuri and Ruslan Salakhutdinov, editors,
  \emph{Proceedings of the 36th International Conference on Machine Learning,
  {ICML} 2019, 9-15 June 2019, Long Beach, California, {USA}}, 2019.

\bibitem[Wang et~al.(2019)Wang, Li, Xiao, Zhu, Li, Wong, and
  Chao]{wang2019learning}
Qiang Wang, Bei Li, Tong Xiao, Jingbo Zhu, Changliang Li, Derek~F. Wong, and
  Lidia~S. Chao.
\newblock Learning deep transformer models for machine translation.
\newblock In \emph{Proceedings of the 57th Annual Meeting of the Association
  for Computational Linguistics}, 2019.

\bibitem[He et~al.(2016)He, Zhang, Ren, and Sun]{he2016deep}
Kaiming He, Xiangyu Zhang, Shaoqing Ren, and Jian Sun.
\newblock Deep residual learning for image recognition.
\newblock In \emph{Proceedings of the IEEE conference on computer vision and
  pattern recognition}, pages 770--778, 2016.

\bibitem[Sennrich et~al.(2016)Sennrich, Haddow, and Birch]{sennrich2015neural}
Rico Sennrich, Barry Haddow, and Alexandra Birch.
\newblock Neural machine translation of rare words with subword units.
\newblock In \emph{Proceedings of the 54th Annual Meeting of the Association
  for Computational Linguistics (Volume 1: Long Papers)}, August 2016.

\bibitem[Baevski and Auli(2019)]{baevski2018adaptive}
Alexei Baevski and Michael Auli.
\newblock Adaptive input representations for neural language modeling.
\newblock In \emph{International Conference on Learning Representations}, 2019.

\bibitem[Grave et~al.(2017{\natexlab{a}})Grave, Joulin, Ciss{\'e}, Grangier,
  and J{\'e}gou]{grave2017efficient}
{\'E}douard Grave, Armand Joulin, Moustapha Ciss{\'e}, David Grangier, and
  Herv{\'e} J{\'e}gou.
\newblock Efficient softmax approximation for {GPU}s.
\newblock In \emph{International Conference on Machine Learning},
  2017{\natexlab{a}}.

\bibitem[Mehta et~al.(2020)Mehta, Koncel-Kedziorski, Rastegari, and
  Hajishirzi]{mehta2020DeFINE}
Sachin Mehta, Rik Koncel-Kedziorski, Mohammad Rastegari, and Hannaneh
  Hajishirzi.
\newblock {DeFINE: Deep Factorized Input Token Embeddings for Neural Sequence
  Modeling}.
\newblock In \emph{International Conference on Learning Representations}, 2020.

\bibitem[Chen et~al.(2018)Chen, Si, Li, Chelba, and Hsieh]{chen2018groupreduce}
Patrick Chen, Si~Si, Yang Li, Ciprian Chelba, and Cho-Jui Hsieh.
\newblock Groupreduce: Block-wise low-rank approximation for neural language
  model shrinking.
\newblock In \emph{Advances in Neural Information Processing Systems}, 2018.

\bibitem[Sun et~al.(2020)Sun, Yu, Song, Liu, Yang, and Zhou]{sun2020mobilebert}
Zhiqing Sun, Hongkun Yu, Xiaodan Song, Renjie Liu, Yiming Yang, and Denny Zhou.
\newblock Mobilebert: a compact task-agnostic bert for resource-limited
  devices.
\newblock In \emph{Association for Computational Linguistics (ACL)}, 2020.

\bibitem[Han et~al.(2016)Han, Mao, and Dally]{han2015deep}
Song Han, Huizi Mao, and William~J Dally.
\newblock Deep compression: Compressing deep neural networks with pruning,
  trained quantization and huffman coding.
\newblock In \emph{International Conference for Representation Learning}, 2016.

\bibitem[Voita et~al.(2019{\natexlab{b}})Voita, Talbot, Moiseev, Sennrich, and
  Titov]{voita2019analyzing}
Elena Voita, David Talbot, Fedor Moiseev, Rico Sennrich, and Ivan Titov.
\newblock Analyzing multi-head self-attention: Specialized heads do the heavy
  lifting, the rest can be pruned.
\newblock In \emph{Proceedings of the 57th Annual Meeting of the Association
  for Computational Linguistics}, 2019{\natexlab{b}}.

\bibitem[Hinton et~al.(2015)Hinton, Vinyals, and Dean]{hinton2015distilling}
Geoffrey Hinton, Oriol Vinyals, and Jeff Dean.
\newblock Distilling the knowledge in a neural network.
\newblock In \emph{NIPS Deep Learning and Representation Learning Workshop},
  2015.

\bibitem[Sanh et~al.(2019)Sanh, Debut, Chaumond, and Wolf]{sanh2019distilbert}
Victor Sanh, Lysandre Debut, Julien Chaumond, and Thomas Wolf.
\newblock Distilbert, a distilled version of bert: smaller, faster, cheaper and
  lighter.
\newblock In \emph{5th Workshop on Energy Efficient Machine Learning and
  Cognitive Computing - NeurIPS}, 2019.

\bibitem[Telgarsky(2016)]{telgarsky2016benefits}
Matus Telgarsky.
\newblock Benefits of depth in neural networks.
\newblock \emph{COLT}, 2016.

\bibitem[Edunov et~al.(2018)Edunov, Ott, Auli, Grangier, and
  Ranzato]{edunov2018classical}
Sergey Edunov, Myle Ott, Michael Auli, David Grangier, and Marc{'}Aurelio
  Ranzato.
\newblock Classical structured prediction losses for sequence to sequence
  learning.
\newblock In \emph{Proceedings of the 2018 Conference of the North {A}merican
  Chapter of the Association for Computational Linguistics: Human Language
  Technologies, Volume 1 (Long Papers)}, 2018.

\bibitem[Vaswani et~al.(2018)Vaswani, Bengio, Brevdo, Chollet, Gomez, Gouws,
  Jones, Kaiser, Kalchbrenner, Parmar, Sepassi, Shazeer, and
  Uszkoreit]{tensor2tensor}
Ashish Vaswani, Samy Bengio, Eugene Brevdo, Francois Chollet, Aidan~N. Gomez,
  Stephan Gouws, Llion Jones, \L{}ukasz Kaiser, Nal Kalchbrenner, Niki Parmar,
  Ryan Sepassi, Noam Shazeer, and Jakob Uszkoreit.
\newblock Tensor2tensor for neural machine translation.
\newblock \emph{CoRR}, abs/1803.07416, 2018.
\newblock URL \url{http://arxiv.org/abs/1803.07416}.

\bibitem[Papineni et~al.(2002)Papineni, Roukos, Ward, and
  Zhu]{papineni2002bleu}
Kishore Papineni, Salim Roukos, Todd Ward, and Wei-Jing Zhu.
\newblock Bleu: a method for automatic evaluation of machine translation.
\newblock In \emph{Proceedings of the 40th annual meeting on association for
  computational linguistics}, pages 311--318. Association for Computational
  Linguistics, 2002.

\bibitem[Ott et~al.(2019)Ott, Edunov, Baevski, Fan, Gross, Ng, Grangier, and
  Auli]{ott2019fairseq}
Myle Ott, Sergey Edunov, Alexei Baevski, Angela Fan, Sam Gross, Nathan Ng,
  David Grangier, and Michael Auli.
\newblock Fairseq: A fast, extensible toolkit for sequence modeling.
\newblock In \emph{Proceedings of NAACL-HLT 2019: Demonstrations}, 2019.

\bibitem[Ghazvininejad et~al.(2019)Ghazvininejad, Levy, Liu, and
  Zettlemoyer]{ghazvininejad2019mask}
Marjan Ghazvininejad, Omer Levy, Yinhan Liu, and Luke Zettlemoyer.
\newblock Mask-predict: Parallel decoding of conditional masked language
  models.
\newblock In \emph{Proceedings of the 2019 Conference on Empirical Methods in
  Natural Language Processing and the 9th International Joint Conference on
  Natural Language Processing (EMNLP-IJCNLP)}, pages 6114--6123, 2019.

\bibitem[Kingma and Ba(2015)]{kingma2014adam}
Diederik~P Kingma and Jimmy Ba.
\newblock Adam: A method for stochastic optimization.
\newblock In \emph{International Conference on Learning Representations}, 2015.

\bibitem[Deng et~al.(2018)Deng, Kim, Chiu, Guo, and Rush]{deng2018latent}
Yuntian Deng, Yoon Kim, Justin Chiu, Demi Guo, and Alexander Rush.
\newblock Latent alignment and variational attention.
\newblock In \emph{Advances in Neural Information Processing Systems}, pages
  9712--9724, 2018.

\bibitem[Merity et~al.(2017)Merity, Xiong, Bradbury, and
  Socher]{merity2017pointer}
Stephen Merity, Caiming Xiong, James Bradbury, and Richard Socher.
\newblock Pointer sentinel mixture models.
\newblock In \emph{International Conference on Learning Representations}, 2017.

\bibitem[Grave et~al.(2017{\natexlab{b}})Grave, Joulin, and
  Usunier]{grave2016improving}
Edouard Grave, Armand Joulin, and Nicolas Usunier.
\newblock Improving neural language models with a continuous cache.
\newblock In \emph{International Conference on Learning Representations},
  2017{\natexlab{b}}.

\bibitem[Merity et~al.(2018{\natexlab{b}})Merity, Keskar, and
  Socher]{merity2018analysis}
Stephen Merity, Nitish~Shirish Keskar, and Richard Socher.
\newblock An analysis of neural language modeling at multiple scales.
\newblock \emph{arXiv preprint arXiv:1803.08240}, 2018{\natexlab{b}}.

\end{thebibliography}
